\documentclass[journal]{IEEEtran}
\usepackage{cite}
\ifCLASSINFOpdf
  \usepackage[pdftex]{graphicx}
\else
  \usepackage[dvips]{graphicx}
\fi
\newtheorem{de}{Definition}
\usepackage{amsmath}
\DeclareMathOperator*{\argmax}{arg\,max}
\DeclareMathOperator*{\argmin}{arg\,min}
\usepackage{amssymb}
\usepackage{amsfonts}
\usepackage{amsopn}
\usepackage{array}
\usepackage{booktabs}
\usepackage[table]{xcolor}
\usepackage[caption=false,font=normalsize,labelfont=sf,textfont=sf]{subfig}
\usepackage[linesnumbered,ruled]{algorithm2e}
\usepackage{url}
\usepackage{microtype}
\usepackage{multirow}
\usepackage{enumerate}
\usepackage{xcolor}

\begin{document}

\title{IGD Indicator-based Evolutionary Algorithm for Many-objective Optimization Problems}

\author{Yanan~Sun,~\IEEEmembership{Member},~\IEEEmembership{IEEE},
        Gary~G.~Yen,~\IEEEmembership{Fellow},~\IEEEmembership{IEEE},
        and~Zhang~Yi,~\IEEEmembership{Fellow},~\IEEEmembership{IEEE}
        \thanks{This work is supported in part by the China Scholarship Council; in part by the Miaozi Project in Science and Technology Innovation Program of Sichuan Province, China; and in part by the National
Natural Science Foundation of China under Grants 61432012 and U1435213.~\emph{(Corresponding author:
Gary G. Yen.)}}
\thanks{Yanan Sun is with the College of Computer Science, Sichuan University, Chengdu 610065 CHINA and with the School of Engineering and Computer Science, Victoria University of Wellington, Wellington 6140 NEW ZEALAND (e-mail:yanan.sun@ecs.vuw.ac.nz).}
\thanks{Gary G. Yen is with the School of Electrical and Computer Engineering, Oklahoma State University, Stillwater, OK 74078 USA (e-mail:gyen@okstate.edu).}
\thanks{Zhang Yi is with the College of Computer Science, Sichuan University, Chengdu 610065 CHINA (e-mail:zhangyi@scu.edu.cn).}
}

\maketitle
\begin{abstract}
Inverted Generational Distance (IGD) has been widely considered as a reliable performance indicator to concurrently quantify the convergence and diversity of multi- and many-objective evolutionary algorithms. In this paper, an IGD indicator-based evolutionary algorithm for solving many-objective optimization problems (MaOPs) has been proposed. Specifically, the IGD indicator is employed in each generation to select the solutions with favorable convergence and diversity. In addition, a computationally efficient dominance comparison method is designed to assign the rank values of solutions along with three newly proposed proximity distance assignments. Based on these two designs, the solutions are selected from a global view by linear assignment mechanism to concern the convergence and diversity simultaneously. In order to facilitate the accuracy of the sampled reference points for the calculation of IGD indicator, we also propose an efficient decomposition-based nadir point estimation method for constructing the Utopian Pareto front which is regarded as the best approximate Pareto front for real-world MaOPs at the early stage of the evolution. To evaluate the performance, a series of experiments is performed on the proposed algorithm against a group of selected state-of-the-art many-objective optimization algorithms over optimization problems with $8$-, $15$-, and $20$-objective. Experimental results measured by the chosen performance metrics indicate that the proposed algorithm is very competitive in addressing MaOPs.
\end{abstract}

\section{Introduction}
\label{section_1}
\IEEEPARstart{M}{any-objective} Optimization Problems (MaOPs) refer to the optimization tasks involving $m$ (i.e., $m>3$) conflicting objectives to be optimized concurrently~\cite{khare2003performance}. Generally, an MaOP is with the mathematic form represented by~(\ref{equ_maop})
\begin{equation}
\label{equ_maop}
\begin{array}{ll}
\setlength{\arraycolsep}{1pt}
\renewcommand{\arraystretch}{1.5}
\left\{
\begin{array}{l}
  \textbf{y}=\textbf{f}(\textbf{x})=(f_1(\textbf{x}),\cdots,f_m(\textbf{x})) \\
  s.t.~~~\textbf{x}\in \Omega
\end{array}
\right.
\end{array}
\end{equation}
where $\Omega \subseteq \mathbb{R}^n$ is the feasible search space for the decision variables $\textbf{x}=(x_1,\cdots,x_n)^T$, and $\textbf{f}: \Omega \rightarrow \Theta\subseteq \mathbb{R}^m$ is the corresponding objective vector including $m$ objectives which maps the $n$-dimensional decision space $\Omega$ to the $m$-dimensional objective space $\Theta$. Without loss of generality, $\textbf{f}(\textbf{x})$ is assumed to be minimized since the maximization problems can be transformed into the minimization problems due to the duality principle. Because of the conflicting nature in the objective functions, there is no single perfect solution for \textbf{f}(\textbf{x}), but a set of tradeoff solutions which form the Pareto Set (PS) in the decision space and the corresponding Pareto Front (PF) in the objective space.

Optimization algorithms for addressing an MaOP aim at searching for a set of uniformly distributed solutions which are closely approximating the PF. Because the MaOPs widely exist in diverse real-world applications, such as policy management in land exploitation with $14$-objective~\cite{chikumbo2012approximating} and calibration of automotive engine with $10$-objective~\cite{lygoe2013real}, to name a few, various algorithms for solving MaOPs have been developed. Among these algorithms, the evolutionary paradigms are considerably preferable due to their population-based meta-heuristic characteristics obtaining a set of quality solutions in a single run.

During the past decades, various Multi-Objective Evolutionary Algorithms (MOEAs), such as elitist Non-dominated Sorting Genetic Algorithm (NSGA-II)~\cite{deb2002fast}, advanced version of Strength Pareto Evolutionary Algorithm (SPEA2)~\cite{zitzler2001spea2}, among others, have been proposed to effectively solve Multi-Objective Optimization Problems~(MOPs). Unfortunately, these MOEAs do not scale well with the increasing number of objectives, mainly due to the loss of selection pressure. To be specific, the number of non-dominated solutions in MaOPs accounts for a large proportion of the current population because of the dominance resistance phenomenon~\cite{fonseca1998multiobjective} caused by the curse of dimensionality~\cite{purshouse2007evolutionary}, so that the traditional elitism mechanism based on Pareto-domination cannot effectively differentiate which solutions should survive into the next generation. As a result, the density-based diversity promotion mechanism is considered the sole mechanism for mating and environmental selections~\cite{li2015many}. However, the solutions with good diversity in MaOPs are generally not only distant from each other but also away from the PF. Consequently, the evolution with the solutions generated by the activated diversity promotion is stagnant or even far away from the PF~\cite{wagner2007pareto}. To this end, various Many-Objective Evolutionary Algorithms (MaOEAs) specifically designed for addressing MaOPs have been proposed in recent years.

Generally, these MaOEAs can be divided into four different categories. The first category covers the algorithms employing reference prior to enhancing the diversity promotion which in turn improve the convergence. For example, the MaOEA using reference-point-based non-dominated sorting approach (NSGA-III)~\cite{deb2014evolutionary} employs a set of reference vectors to assist the algorithm to select solutions which are close to these reference vectors. Yuan \textit{et al.}~\cite{yuan2016new} proposed the reference line-based algorithm which not only adopted the diversity improvement mechanism like that in NSGA-III but also introduced convergence enhancement scheme by measuring the distance between the origin to the solution projections on the corresponding reference line. In addition, a reference line-based estimation of distribution algorithm was introduced in~\cite{sun2017reference} for explicitly promoting the diversity of an MaOEA. Furthermore, an approach (RVEA) was presented in~\cite{cheng2016reference} to adaptively revise the reference vector positions based on the scales of the objective functions to balance the diversity and convergence.

The second category refers to the decomposition-based algorithms which decompose an MaOP into several single-objective optimization problems, such as the MOEA based on Decomposition (MOEA/D)~\cite{zhang2007moea} which was initially proposed for solving MOPs but scaled well for MaOPs. Specifically, MOEA/D transformed the original MOP/MaOP with $m$ objectives into a group of single-objective optimization problems, and each sub-problem was solved in its neighboring region which constrained by their corresponding reference vectors. Recently, diverse variants~\cite{yuan2016balancing, wang2014replacement,li2014stable, li2015interrelationship, asafuddoula2015decomposition, gee2015online} of MOEA/D were proposed for improving the performance much further.

The third category is known as the convergence enhancement-based approaches. More specifically, the traditional Pareto dominance comparison methods widely utilized in MOEAs are not effective in discriminating populations with good proximity in MaOPs. A natural way is to modify this comparison principle to promote the selection mechanism. For example, the $\epsilon$-dominance method~\cite{laumanns2002combining} employed a relaxed factor $\epsilon$ to compare the dominance relation between solutions; Pierro \textit{et al}.~\cite{di2007investigation} proposed the preference order ranking approach to replace the traditional non-dominated sorting. Furthermore, the fuzzy dominance methods~\cite{wang2007fuzzy,he2014fuzzy} studied the fuzzification of the Pareto-dominance relation to design the ranking scheme to select promising solutions; the $L$-optimality paradigm was proposed in~\cite{zou2008new} to pick up solutions whose objectives were with the same importance by considering their objective value improvements. In addition, Yang \textit{et al.}~\cite{yang2013grid} proposed the grid-based approach to select the solutions that have the higher priority of dominance, and control the proportion of Pareto-optimal solutions by adjusting the grid size. Meanwhile, Antonio \textit{et al.}~\cite{l2013alternative} alternated the achievement function and the $\epsilon$-indicator method to improve the performance of MOEA in solving MaOPs. In~\cite{li2014shift}, a modification of density estimation, termed as shift-based density estimation, was proposed to make the dominance comparison better suited for solving MaOPs. Furthermore, the favorable convergence scheme was proposed in~\cite{cheng2015many} to improve the selection pressure in mating and environmental selections. Recently, a knee point-based algorithm (KnEA)~\cite{zhang2015knee} was presented as a secondary selection scheme to enhance the selection pressure. In summary, these algorithms introduced new comparison methods, designed effective selection mechanisms, or relaxed the original comparison approach to improve the selection pressure in addressing MaOPs.

The fourth category is known as the indicator-based methods. For instance, several MOEAs based on the hypervolume (HV) were proposed in~\cite{emmerich2005emo,igel2007covariance,brockhoff2007improving}, however their major disadvantages were the costly overhead in calculating the HV values especially in solving MaOPs. To this end, Bader and Zitzler proposed the HypE method with the Monte Carlo simulation~\cite{bader2011hype} to estimate the HV value. Consequently, the computational cost was largely lowered compared to its predecessors whose HV values were calculated exactly. In~\cite{gerstl2011finding}, an $\bigtriangleup p$ indicator-based algorithm ($\bigtriangleup p$-EMOA) was proposed for solving bi-objective optimization problems, and then extended further for tri-objective problems~\cite{trautmann2013finding}. Furthermore, Villalobos and Coello~\cite{rodriguez2012new} integrated the $\bigtriangleup p$ indicator with the differential evolution~\cite{storn1995differential} to solve MaOPs with up to $10$ objectives. Recently, an Inverse Generational Distance Plus (IGD$^+$)~\cite{ishibuchi2015modified} indicator-based evolutionary algorithm (IGD$^+$-EMOA) was proposed in~\cite{lopez2016igd+} for addressing MaOPs with no more than $8$ objectives. Basically, the IGD$^+$ indicator is viewed as a variant of the Inverse Generational Distance (IGD) indicator.

Although the MaOEAs mentioned above have experimentally demonstrated their promising performance, major issues are easily to be identified in solving real-world applications. For example, it is difficult to choose the converting strategy of the MaOEAs from the second category, which motivates multiple variants~\cite{yuan2016balancing,sun2016manifold,li2014stable,sun2017global,asafuddoula2015decomposition, gee2015online} to be developed further. In addition, the MaOEAs from the first and third categories only highlighted one of the characters in their designs (i.e., only the diversity promotion is explicitly concerned in the MaOEAs from the first category, and the convergence from the third category). However, both the diversity and convergence are concurrently desired by the MaOEAs. In this regard, some performance indicators, which are capable of simultaneously measuring the diversity and convergence, such as the HV and IGD indicators, are preferred to be employed for designing MaOEAs. However, the major issue of HV is its high computational complexity. Although the Monte Carlo simulation has been employed to mitigate this adverse impact, the calculation is still impracticable when the number of objectives is more than 10~\cite{bader2011hype}, while the calculation of IGD is scalable without these deficiencies.

In this paper, an IGD indicator-based Many-Objective Evolutionary Algorithm (MaOEA/IGD) has been proposed for effectively addressing MaOPs, and the contributions are outlined as follows:

\begin{enumerate}
 \item A Decomposition-based Nadir Point Estimation method (DNPE) has been presented to estimate the nadir points to facilitate the calculation of IGD indicator. In DNPE, the estimation focuses only on the extreme point areas and transforms the computation of an $m$-objective optimization problem into $m$ single-objective optimization problems. Therefore, less computational cost is required compared to its peer competitors (experiments are demonstrated in Subsection~\ref{section_experiment_nadir}).
  \item A comparison scheme for the non-dominated sorting has been designed for improving the convergence of the proposed algorithm. In this scheme, the dominance relations of solutions are not obtained by the comparisons among all the solutions but the solutions to the reference points. Therefore, the computational complexity of the presented comparison scheme is significantly lessened compared to the traditional comparison means because the number of reference points is generally much less than that of the whole population.
  \item Three types of proximity distance assignment mechanisms are proposed for the solutions according to their PF rank values, which make the solutions with good convergence in the same PF to have higher chances to be selected. Furthermore, these assignment mechanisms collectively assure the proposed IGD indicator to be Pareto compliance.
  \item Based on the proposed dominance comparison scheme and the proximity distance assignments, the selection mechanism which is employed for the mating selection and the environmental selection is proposed to concurrently facilitate the convergence and the diversity.
\end{enumerate}

The reminder of this paper is organized as follows. First, related works are reviewed, and the motivation of the proposed DNPE is presented in Section~\ref{section_2}. Then the details of the proposed algorithm are documented in Section~\ref{section_3}. To evaluate the performance of the proposed algorithm in addressing MaOPs, a series of experiments over scalable benchmark test suits are performed against state-of-the-art MaOEAs, and their results are measured by commonly chosen performance metrics and then analyzed in Section~\ref{section_4}. In addition, the performance of the proposed MaOEA/IGD is also demonstrated by solving a real-world application, and the performance of the proposed DNPE in nadir point estimation is investigated against its peer competitors. Finally, the proposed algorithm is concluded and the future works are illustrated in Section~\ref{section_5}.

\section{Related Works and Motivation}
\label{section_2}
Literatures related to the nadir point estimation and IGD indicator-based EAs are thoroughly reviewed in this section. Specifically, the Worst Crowded NSGA-II (WC-NSGA-II)~\cite{deb2006towards} and the Pareto Corner Search Evolutionary Algorithm (PCSEA)~\cite{singh2011pareto} would be reviewed and criticized in detail, because the insightful observations of the deficiencies of these two approaches naturally lead to the motivation of the proposed DNPE design. In addition, the IGD$^+$-EMOA is reviewed as well to highlight the utilization of the IGD$^+$ indicator and the reference points sampling for the calculation of IGD in the proposed MaOEA/IGD. Please note that all the discussions in this section are with the context of the problem formulation in~(\ref{equ_maop}).
\subsection{Nadir Point Estimation Methods}
\label{section_2_1}

According to literatures~\cite{wang2015nadir,deb2008review}, the approaches for estimating the nadir points can be divided into three categories including the surface-to-nadir, edge-to-nadir, and extreme-point-to-nadir schemes. In the surface-to-nadir scheme, the nadir points are constructed from the current Pareto-optimal solutions, and updated as the corresponding algorithms evolve towards the PF. MOEAs in ~\cite{deb2002fast,ke2013moea,ma2016multiobjective,chen2015evolutionary,li2015interrelationship} and MaOEAs in~\cite{deb2014evolutionary,yuan2015balancing,cheng2016reference} belong to this category. However, these MOEAs are shown to perform poorly in MaOPs due to the curse of dimensionality~\cite{praditwong2007well}. In addition, the MaOEAs related methods are not suitable for the proposed algorithm because the MaOPs have been solved prior to the nadir point estimation, while the nadir points in this paper are targeted for addressing MaOPs.

The edge-to-nadir scheme covers the Marcin and Andrzej's approach~\cite{szczepanski2003application}, Extremized Crowded NSGA-II (EC-NSGA-II)~\cite{deb2006towards}, and the recently proposed Emphasized Critical Region (ECR) approach~\cite{wang2015nadir}. Specifically, Marcin and Andrzej's approach decomposed an $m$-objective problem into $C^2_m$ sub-problems to estimate the nadir point from the $C^2_m$ edges, in which the major issues were the poor quality in nadir point found and the impractical computation complexity beyond three objectives~\cite{wang2015nadir}. EC-NSGA-II modified the crowding distance of NSGA-II by assigning large rank values to the solutions which had the minimum or maximum objective values. The ECR emphasized the solutions lying at the edges of the PF (i.e., the critical regions) with the adopted MOEAs. Although EC-NSGA-II and ECR have been reported to be capable of estimating the nadir points in MaOPs, they required a significantly large number of functional evaluations~\cite{wang2015nadir}.

The extreme-point-to-nadir approaches refer to employ a direct means to estimate the extreme points based on which the nadir points are derived, such as the Worst Crowded NSGA-II (WC-NSGA-II)~\cite{deb2006towards} in which the worst crowded solutions (extreme points) were preferred by ranking their crowding distances with large values. In WC-NSGA-II, it was hopeful that the extreme points were obtained when the evolution terminated. However, emphasizing the extreme points easily led to the WC-NSGA-II losing the diversity which inadvertently affected the convergence in turn. In addition, Singh \textit{et al.} proposed the Pareto Corner Search Evolutionary Algorithm (PCSEA)~\cite{singh2011pareto} to look for the nadir points with the corner-sort ranking method for the MaOPs whose objective values were required to be with the identical scales.

In addition, there are also various methods not falling into the above categories. For example, Benayoun \textit{et al.} estimated the nadir points with the pay-table~\cite{benayoun1971linear} in which the $j$-th row denoted the objective values of the solution which had the minima on its $j$-th objective. In addition, other related works were suggested in~\cite{{dessouky1986estimates,isermann1988computational,korhonen1997heuristic}} for the problems assuming a linear relationship between the objectives and variables. On the contrary, most of the real-world applications are non-linear in nature.

Because the nadir point estimation is a critical part of the proposed algorithm for solving MaOPs, an approach with a high computational complexity is certainly not preferable. Furthermore, the nadir points are employed for constructing the Utopian PF, while the reference point of IGD would come from the PF. As a consequence, nadir points with a high accuracy are not necessary a guarantee. Considering the balance between the computational complexity and the estimation accuracy, we have proposed in this paper a Decomposition-based Nadir Point Estimation method (DNPE) by transforming an $m$-objective MaOP into $m$ single-objective optimization problems to search for the respective $m$ extreme points. Thereafter the nadir point is derived.

Specifically, because the proposed nadir point estimation method (i.e., the DNPE) is based on the extreme-point-to-nadir scheme, the WC-NSGA-II and the PCSEA which are with the similar scheme are detailed further. For convenience of reviewing the related nadir point estimation methods, multiple fundamental concepts of the MaOPs are given first. Then the WC-NSGA-II and PCSEA are discussed.
\begin{figure}[htp]
  \centering
  \includegraphics[width=0.8\columnwidth]{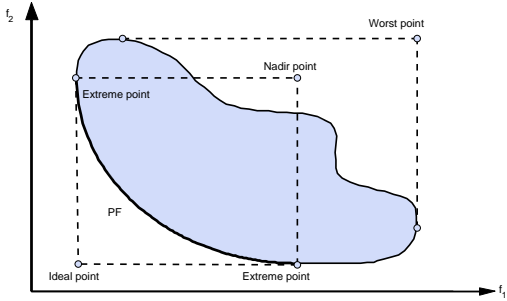}\\
  \caption{An example with bi-objective optimization problem to illustrate the ideal point, extreme point, worst point, nadir point, and the PF.}\label{fig_all_points}
\end{figure}

\begin{de}
Generally, there are $m$ extreme points denoted as $\textbf{y}^{ext}_1$, $\cdots$, $\textbf{y}^{ext}_m$ in an $m$-objective optimization problem,
$\textbf{y}^{ext}_i=\textbf{f}(\textbf{x}^{ext}_i)$, and $\textbf{x}^{ext}_i = \argmax_{\textbf{x}}~f_i(\textbf{x}),$ where $\textbf{x}\in \text{PS}$ and $i \in \{1,\cdots,m\}$.
\end{de}

\begin{de}
\label{definition_nadir_point}
The nadir point is defined as $\textbf{z}^{nad}=\{z_1^{nad}, \cdots, z_m^{nad}\}$, where $z_i^{nad}=f_i(\textbf{x}^{ext}_i)$.
\end{de}
\begin{de}
The worst point is defined as $\textbf{z}^{w}=\{z_1^{w},\cdots,z_m^{w}\}$, where $z_i^w = \text{max}~f_i(\textbf{x})$ and $\textbf{x} \in \Omega$.
\end{de}
\begin{de}
\label{definition_ideal_point}
The ideal point is defined as $\textbf{z}^*=\{z_1^*, \cdots, z_m^*\}$, where $z_i^*=\text{min}~f_i(\textbf{x})$ and $\textbf{x} \in \Omega$.
\end{de}

Furthermore, the ideal point, extreme point, worst point, nadir point, and the PF are plotted with a bi-objective optimization problem in Fig.~\ref{fig_all_points} for intuitively understanding their significance. With these fundamental definitions, a couple of nadir point estimation algorithms, WC-NSGA-II and PCSEA, which are in the extreme-point-to-point scheme are discussed as follows.

 WC-NSGA-II was designed based on NSGA-II by modifying its crowding distance assignment. According to the definition of nadir point in Definition~\ref{definition_nadir_point}, WC-NSGA-II naturally emphasized the solutions with maximal objectives front-wise. Specifically, solutions on a particular non-dominated front were sorted with an increasing order based on their fitness, and rank values equal to their positions in the ordered list were assigned. Then the solutions with larger rank values were preferred in each generation during the evolution. By this emphasis mechanism, it was hopeful that nadir point was obtained when the evolution of WC-NSGA-II was terminated. However, one major deficiency is that over-emphasis on these solutions with maximal fitness leads to the lack of diversity, which in turn affects the convergence of the generated extreme points, i.e., the generated extreme points in WC-NSGA-II are not necessarily Pareto-optimal.

\begin{figure}
  \centering
  \includegraphics[width=0.8\columnwidth]{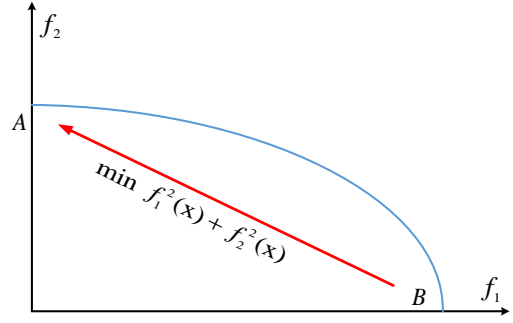}\\
  \caption{A bi-objective example to illustrate the deficiency of Pareto Corner Search Evolutionary Algorithm in addressing the problem with different objective value scales.}\label{fig_pcsea_problem_1}
\end{figure}

PCSEA employed the corner-sorting to focus on the extreme points during the evolution. Specifically, there were $2m$ ascended lists during the executions of PCSEA. The first $m$ lists were about the $m$ objectives of the solutions, while the other $m$ lists were about the excluded square $L_2$ norm with each objective. Furthermore, the $j$-th objective of the problem to be optimized was with the excluded square $L_2$ norm $\sum_{i=1, i\neq j}^{m}f_i (\textbf{x})^2$. From these $2m$ lists, solutions with smaller rank values which were equal to their positions in these lists were selected until there was no available slot. Experimental results have shown that PCSEA performs well in MaOPs due to the utilization of corner-sorting other than the non-dominated sorting which easily leads to the loss of selection pressure. However, the corner-sorting can be viewed as to minimize the square $L_2$ norm of all objectives, which deteriorates the performance of PCSEA in solving the problems with different objective value scales and non-concave PF shapes. For example in Fig.~\ref{fig_pcsea_problem_1} which illustrates an example of bi-objective optimization problem, the arc $AB$ denotes the PF, points $A$ and $B$ are with different values. It is clearly observed that if the minimization of $L_2$ norm regarding $f_1({\textbf{x}})$ and $f_2({\textbf{x}})$ are emphasized, only the extreme point $A$ would be obtained while the other one (point $B$) would be missed. This deficiency can also be seen in the problems with non-concave PF shapes.

\begin{figure}
  \centering
  \includegraphics[width=0.8\columnwidth]{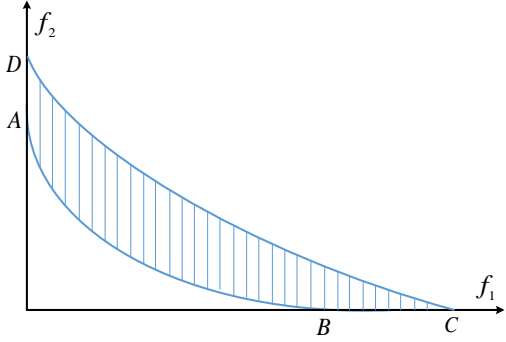}\\
  \caption{An example of bi-objective optimization problem, $A,B$ are the extreme points, and $C,D$ are the worse points. The shade region denotes the feasible objective space.}\label{fig_motivation_extreme_point}
\end{figure}

Briefly, major concerns in these two nadir point estimation algorithms are summarized as 1) over-emphasizing extreme points leads to the loss of diversity which in turn deteriorates the convergence of the found nadir points, and 2) simultaneously minimizing the objectives does not scale to problems with different objective value scales and non-concave PF shapes. To this end, a natural approach is recommended by 1) decomposing the problem to be solved into several single-objective optimization problems in which the diversity is not required, and 2) assigning different weights to the objectives. In the proposed DNPE, the $m$ respective extreme points are estimated by decomposing the $m$-objective MaOP into $m$ single-objective problems associated with different weights. Specifically, the $i$-th extreme point estimation is with the form formulated by~(\ref{equ_extrme_point})
\begin{equation}
\label{equ_extrme_point}
\text{min}~|f_i(\textbf{x})| + \lambda \sum_{j=1, j\neq i}^{m}(f_j(\textbf{x}))^2
\end{equation}
where $\lambda$ is a factor with the value greater than $1$ to highlight the priority of solving its associated term. In order to better justify our motivation, an example with bi-objective is plotted in Fig.~\ref{fig_motivation_extreme_point} in which $A,B$ are the extreme points, $C,D$ are the worse points, and the shaded region denotes the feasible objective space. To obtain the extreme point on the $f_1$ objective, it is required to minimize $|f_1(\textbf{x})| + \lambda (f_2(\textbf{x}))^2$ according to~(\ref{equ_extrme_point}). Because $\lambda$ is greater than $1$, the term ${(f_2(\textbf{x}))^2}$ is optimized with a higher priority. Consequently, solutions locating in line $BC$ are obtained, based on which $|f_1(\textbf{x})|$ is minimized much further, then the extreme point $B$ is obtained.

\subsection{IGD$^{+}$-EMOA}
Prior to the introduction of IGD$^{+}$-EMOA, it is necessary to compare the differences between the IGD and IGD$^+$ indicators. For this purpose, we first list their respective mathematical formulations. Then the superiority of IGD$^+$ indicator is highlighted. Finally, the IGD$^+$-EMOA is discussed much further.

Basically, the IGD indicator is with the form formulated by~(\ref{equ_igd_indicator})
\begin{equation}
\label{equ_igd_indicator}
\mathrm{IGD} = \frac{\sum_{p \in p^*}\text{dist}(p,PF)}{|p^*|},
\end{equation}
where $p^*$ denotes a set of reference points in the calculation of IGD, $PF$ denotes the non-dominated solutions generated by the algorithm, $dist(p,PF)$ denotes the nearest distance from $p$ to solutions in $PF$, and the distance from $p$ to the solution $y$ in $PF$ is calculated by $d(p,y) = \sqrt{\sum_{j=1}^m(p_j-y_j)^2}$. It has been pointed out in~\cite{ishibuchi2015modified} that IGD cannot differentiate the quality of generated solutions when they are non-dominated to the solutions in $p^*$, and the IGD$^+$ indicator is proposed by changing the calculation of $d(p,y)$ to $\sqrt{\sum_{j=1}^m \text{max}(y_j-p_j, 0)^2}$.

IGD$^+$-EMOA employed the IGD$^+$ indicator as its selection mechanism. In addition, the $p^*$ in IGD$^+$-EMOA is sampled from the approximate PF. Specifically, it supposed that the PF is obtained by solving $y_1^r+\cdots+y_m^r=1$ where $\textbf{y}=\{y_1,\cdots,y_m\}$ is from the non-dominated solutions of the current population. However, this approximate approach for generating the PF performs badly in MaOPs where multiple local Pareto-optimal exist, which leads IGD$^+$-EMOA only to solve MaOPs with no more than $8$ objectives~\cite{lopez2016igd+}.

In summary, we first introduce details of the proposed DNPE which is motivated by the insightful observations in deficiencies of WC-NSGA-II and PCSEA. With the help of the estimated nadir point, the Utopian PF is constructed and the reference points are sampled for the calculation of the proposed MaOEA/IGD. Compared to the approximation of the PF in IGD$^+$-EMOA, our proposed MaOEA/IGD is capable of solving problems with many more objectives. Considering the superiority of IGD$+$, its design principle is employed in the proposed MaOEA/IGD when the generated solutions are non-dominated to the sampled reference points.

\section{Proposed algorithm}
\label{section_3}
In this section, the proposed Inverted Generational Distance indicator-based evolutionary algorithm for addressing many-objective optimization problems (in short for MaOEA/IGD) is presented. To be specific, the framework of the proposed algorithm is outlined first. Then the details of each step in the framework are documented. Next, the computational complexity of the proposed algorithm is analyzed. Finally, the mechanisms of promoting the diversity and the convergence in the proposed algorithm are discussed. Noted here that, the proposed algorithm is described within the context formulated by~(\ref{equ_maop}).

\subsection{Framework of the Proposed Algorithm}
\label{section_3_1}
Because the proposed algorithm is based on the IGD indicator, a set of uniformly distributed points which is generated from the PF is required. However, exact points are difficult to be obtained due to the unknown analytical form of the PF for a given real-world application. In the proposed algorithm, a set of reference points, denoted by $p^*$, which are evenly distributed in the Utopian PF is generated first (Subsection~\ref{section_3_2}). Then the population with the predefined size is randomly initialized in the feasible space, and their fitness are evaluated. Next, the population begins to evolve in pursuing the optima until the stopping conditions are satisfied. When the evolution terminates, a number of promising solutions which are hopeful to uniformly distributed in the PF with good proximity is obtained. In order to remedy the shortage caused by using the $p^*$ as the Pareto-optimal solutions with promising diversity for IGD indicator, the rank of each solution as well as its proximity distances to all the points in $p^*$ are assigned (Subsection~\ref{section_3_3}) first during each generation of the evolution. Then new offspring are generated from their parents selected based on the comparisons over their corresponding rank values and proximity distances (Subsection~\ref{section_3_4}). Next, the fitness, ranks, and the proximity distances of the generated offspring to the solutions in $p^*$ are calculated. Finally, a limit number of individuals is selected from the current population to survive into the next generation with the operation of environmental selection (Subsection~\ref{section_3_5}). In summary, the details of the framework are listed in Algorithm~\ref{alg_the_proposed_algorithm}.

\begin{algorithm}
  \caption{Framework of the Proposed Algorithm}
  \label{alg_the_proposed_algorithm}
    $p^* \leftarrow$ Uniformly generate reference points for IGD indicator;\\
    \label{alg_framework_line_1}
    $P_0 \leftarrow$ Randomly initialize the population;\\
    \label{alg_framework_line_2}
    Fitness evaluation on $P_0$;\\
    \label{alg_framework_line_3}
    $t\leftarrow 0$;\\
    \While{stopping criteria are not satisfied}
    {\label{alg_framework_start_while}
        Assign the ranks and the proximity distances for individuals in $P_t$;\\
        \label{alg_framework_line_4}
        $Q_t \leftarrow$ Generate offspring from $P_t$;\\
        \label{alg_framework_line_5}
        Fitness evaluation on $Q_t$;\\
        \label{alg_framework_line_6}
        Assign the rank and the proximity distance for each individual in $ Q_t$;\\
        \label{alg_framework_line_7}
        $P_{t+1}\leftarrow$ Environmental selection from $Q_t\cup P_t$;\\
        \label{alg_framework_line_8}
        $t\leftarrow t+1$;\\
    }
    \label{alg_framework_end_while}
    \textbf{Return} $P_{t}$.
\end{algorithm}

\subsection{Uniformly Generating Reference Points}
\label{section_3_2}
In order to obtain the $p^*$, the extreme points of the problem $\textbf{f}(\textbf{x})$ to be optimized are calculated first. Then the ideal point and the nadir point are extracted. Next, a set of solutions is uniformly sampled from the constrained $(m-1)$-dimensional hyperplane. Finally, these solutions are transformed into the Utopian PF for the usage of the IGD indicator based on the ideal point and the nadir point. Furthermore, these steps are listed in Algorithm~\ref{alg_obtain_reference_points}, and all the details of obtaining the $p^*$ are illustrated as follows.

\begin{algorithm}
  \caption{Uniformly Generate $p^*$ for IGD Indicator}
  \label{alg_obtain_reference_points}
    \KwIn{Optimization problem $\textbf{f}(\textbf{x})=(f_1(\textbf{x}),\cdots,f_m(\textbf{x}))$; the size $k$ of $p^*$.}
    \KwOut{$p^*$.}
     Estimate the extreme points of $\textbf{f}(\textbf{x})$ with Algorithm~\ref{alg_estimate_extreme_points};\\
    $\textbf{z*}=\{z_1^*, \cdots, z_m^*\} \leftarrow$ Extract the ideal point;\\
    $\textbf{z}^{nad}=\{z_1^{nad}, \cdots, z_m^{nad}\} \leftarrow$ Extract the nadir point;\\
    $p^* \leftarrow$ Uniformly generate $k$ points from the constrained hyperplane;\\
    \label{alg_generate_pf_in_unit_hyperplane}
    \For{$i\leftarrow 1$ \rm{\textbf{to}} $k$}{
    \label{alg_begin_to_transform_pf}
        \For{$j\leftarrow$ \rm{\textbf{to}} $m$} {
            $(p^*)^i_j$ = $(p^*)^i_j\times (z_j^{nad}-z_j^*) + z_j^*$
        }
    }
    \label{alg_end_to_transform_pf}
    \textbf{Return} $p^*$.
\end{algorithm}

To estimate the extreme points, the motivation mentioned in Subsection~\ref{section_2_1} is implemented, and the details are presented in Algorithm~\ref{alg_estimate_extreme_points}. Specifically, the $m$ extreme points are estimated individually based on the $m$ objectives of the optimization problem. Furthermore, to estimate the $i$-th extreme point $\textbf{y}_i^{ext}$, the square $L_2$ norm of $\{f_k(\textbf{x})|k = 1,\cdots, i-1, i+1,\cdots, m\}$ is calculated first. Then the absolute value of $f_i(\textbf{x})$ is calculated. Mathematically, these two steps are formulated by $f_{l2}(\textbf{x})=\sum_{k=1,k\neq i}^m||f_k(\textbf{x})||_2^2$ and $f_{l1}(\textbf{x})=|f_i(\textbf{x})|$, respectively, where $\|\cdot\|_{2}$ is the $L_2$ norm operator, and $|\cdot|$ is the absolute value operator. Finally, the extreme point $\textbf{y}_i^{ext}$ is obtained by line~\ref{alg_estimate_extreme_points_optimization_step} in Algorithm~\ref{alg_estimate_extreme_points}, where $\lambda$ is a factor with the value greater than $1$ to highlight the weight of the corresponding term in the optimization. When all the extreme points have been estimated, the nadir point and the ideal point are extracted from the extreme points based on Definitions~\ref{definition_nadir_point} and~\ref{definition_ideal_point}, respectively. This is followed by generating a set of uniformly distributed reference points denoted by $p^*$ from the $(m-1)$-dimensional constrained hyperplane which is contoured by $m$ lines with the unit intercepts in the positive part of the quadrant. Noted here that the Das and Dennis's method~\cite{das1998normal}, which is widely used by some state-of-the-art MaOEAs, such as MOEA/D~\cite{zhang2007moea} and NSGA-III~\cite{deb2014evolutionary}, is employed for the generation of $p^*$ (line~\ref{alg_generate_pf_in_unit_hyperplane} of Algorithm~\ref{alg_obtain_reference_points}). Ultimately, all the points in $p^*$ are transformed into the Utopian PF, which are detailed in lines~\ref{alg_begin_to_transform_pf}-\ref{alg_end_to_transform_pf} of Algorithm~\ref{alg_obtain_reference_points}.

\begin{algorithm}
  \caption{Estimate Extreme Points}
  \label{alg_estimate_extreme_points}
    \KwIn{Optimization problem $\textbf{f}(\textbf{x})=(f_1(\textbf{x}),\cdots,f_m(\textbf{x}))$.}
    \KwOut{Extreme points $\{\textbf{y}_1^{ext}, \cdots, \textbf{y}_m^{ext}\}$.}
    $\Upsilon\leftarrow\emptyset$;\\
    \For{$i\leftarrow 1$ \rm{\textbf{to}} $m$}{
        $\Gamma\leftarrow\emptyset$;\\
        \For{$k\leftarrow 1$ \rm{\textbf{to}} $m$}{
            \If{$k\neq i$}{
                $\Gamma\leftarrow$ $\Gamma\cup f_k(\textbf{x})$;
            }
        }
        $\textbf{x}_i^{ext} = \argmin_{\textbf{x}}~\lambda \|\Gamma\|_2^2 + |f_i(\textbf{x})|$;\\

        $\textbf{y}_i^{ext}\leftarrow f_i(\textbf{x}_i^{ext})$;\\
        \label{alg_estimate_extreme_points_optimization_step}
        $\Upsilon\leftarrow$ $\Upsilon\cup \textbf{y}_i^{ext}$;
    }
    \textbf{Return} $\Upsilon$.
\end{algorithm}

\subsection{Assigning Ranks and Proximity Distances}
\label{section_3_3}
When $p^*$ has been generated, the population is randomly initialized in the feasible search space first, and then the fitness of individuals are evaluated. Next, the rank value and the proximity distances of each individual are assigned. It is noted here that, the rank values are used to distinguish the proximity of the solutions to the Utopian PF from the view of reference points, and the proximity distances are utilized to indicate which individuals are with better convergence and diversity in the sub-population in which the solutions are with the same rank values. More details are discussed in Subsection~\ref{section_3_7}.

Particularly, three rank values, denoted by $r_1$, $r_2$, and $r_3$, exist in the proposed algorithm for all the individuals. Specifically, the way to rank individual $s$ is based on the definitions given as follows.
\begin{de}
\label{definition_r1}
  Individual $s$ is ranked as $r_1$, if it dominates at least one solution in $p^*$.
\end{de}
\begin{de}
\label{definition_r2}
  Individual $s$ is ranked as $r_2$, if it is non-dominated to all the solutions in $p^*$.
\end{de}
\begin{de}
\label{definition_r3}
  Individual $s$ is ranked as $r_3$, if it is dominated by all the solutions in $p^*$, or dominated by a part of solutions in $p^*$ but non-dominated to the remaining solutions.
\end{de}

With Definitions~\ref{definition_r1}, \ref{definition_r2}, and \ref{definition_r3}, it is concluded that Pareto-optimal solutions are all with rank values $r_1$, $r_2$, and $r_3$, if the PF is convex, a hyperplane, and concave, respectively\footnote{This is considered in the context of a minimization problem with continuous PF, and the extreme points are excluded from the Pareto-optimal solutions.}. To be specific, if the PF of a minimization problem is a hyperplane, the Utopian PF is obviously equivalent to the PF. Consequently, the Pareto-optimal solutions lying at the PF are all non-dominated to the reference points which are sampled from the Utopian PF. Based on Definition~\ref{definition_r2}, the Pareto-optimal solutions are ranked with $r_2$. This is also held true for the Pareto-optimal solutions ranked with $r_1$ for convex PF and Pareto-optimal solutions ranked with $r_3$ for concave PF.

Based on the conclusion mentioned above, the proximity distances of each individual with different ranks in the population are calculated. For convenience, it is assumed that there are $k$ solutions in $p^*$, and $q$ individuals in the current population. Consequently, there will be $q\times k$ proximity distances. Let $d^i_j$ denote the proximity distance of $\textbf{f}(\textbf{x}^i)=(f_1(\textbf{x}^i),\cdots,f_m(\textbf{x}^i))$ ($\textbf{x}^i$ refers to the $i$-th individual) to the $j$-th point in $(\textbf{p}^*)^j=\left((p^*)^j_1,\cdots,(p^*)^j_m\right)$, where $i=\{1,\cdots,q\}$ and $j=\{1,\cdots, k\}$. Due to one individual with multiple proximity distances, the corresponding minimal proximity distance would be employed when two individuals are compared upon their proximity distances. Because the proximity distance assignment is used to differentiate the convergence of individuals with the same rank values by comparing their associate proximity distances when a prior knowledge of the PF is unknown in advance, the rank $r^i$ of $\textbf{x}^i$ is confirmed first in order to calculate the $d^i_j$. Particularly, the proximity distance assignment is designed as follows. If $r^i$ is equal to $r_3$, $d^i_j$ is set to the Euclidean distance between $\textbf{f}(\textbf{x}^i)$ and $(\textbf{p}^*)^j$; if $r^i$ is equal to $r_1$, $d^i_j$ is set to the negative value of the Euclidean distance between $\textbf{f}(\textbf{x}^i)$ and $(\textbf{p}^*)^j$; if $r^i$ is equal to $r_2$, $d^i_j$ is calculated by~(\ref{equ_calculate_proximity_distance_r2}).
\begin{equation}
\label{equ_calculate_proximity_distance_r2}
  d^i_j = \sqrt{\sum_{l=1}^m \text{max}(f_l(\textbf{x}^i)-(p^*)^j_l, 0)^2}
\end{equation}

\begin{figure}
  \centering
  \includegraphics[width=0.7\columnwidth]{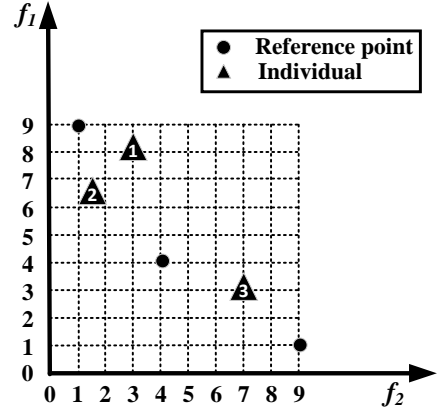}\\
  \caption{A bi-objective optimization problem is illustrated to show the motivation of the proximity distance assignment for the individuals with rank value $r_2$.}\label{fig_igd_plus_comparison}
\end{figure}

For an intuitive understanding, an example is illustrated in Fig.~\ref{fig_igd_plus_comparison} to present the motivation of the proximity distance assignment for individuals who are ranked as $r_2$. In Fig.~\ref{fig_igd_plus_comparison}, the black solid circles refer to the reference points, the black triangles marked by $1$, $2$, and $3$ refer to the individuals $\textbf{x}^1$, $\textbf{x}^2$, and $\textbf{x}^3$ which are with the rank $r_2$ (these three individuals are non-dominated to the reference points). In this situation, it is clearly shown that the individual $\textbf{x}^1$ is with the smallest minimal proximity distance if the Euclidean distance metric is employed (the minimal proximity distances of individuals $\textbf{x}^1$, $\textbf{x}^2$, and $\textbf{x}^3$ are $2.2361$, $2.5495$, and $2.8284$, respectively). Consequently, both the distance measurements of the individuals with ranks $r_1$ and $r_3$ in this situation cannot be utilized to select the desirable individual $\textbf{x}^2$ which has the most promising convergence to the PF.
 However, if the proximity distance quantified by~(\ref{equ_calculate_proximity_distance_r2}) is employed, it is clearly that the individual $\textbf{x}^2$ is with the smallest minimal proximity distance (the minimal proximity distances of individuals $\textbf{x}^1$, $\textbf{x}^2$, and $\textbf{x}^3$ are $2$, $0.5$, and $2$, respectively), which satisfies the motivation of the proximity distance assignment that a smaller value implies a better convergence.
Noted here that, the proximity distance assignment for the individuals with rank $r_2$ is also employed in the IGD$^+$ indicator~\cite{ishibuchi2015modified}.
In summary, smaller proximity distance reveals the better proximity when the exact PF of the problem to be optimized is unknown. Furthermore, the algorithm of the proximity distance assignment is presented in Algorithm~\ref{alg_assign_proximity_distance}.

\begin{algorithm}
  \caption{Assign Proximity Distance}
  \label{alg_assign_proximity_distance}
    \KwIn{Current population $P_t$ with size $q$; reference points $p^*$ with size $k$.}
    \KwOut{Proximity distances matrix $d$.}
    \For{$i\leftarrow1$ \rm{\textbf{to}} $q$}{
        $\textbf{x}^i\leftarrow$ $P_t^i$;\\
        $r\leftarrow$ Calculate the rank of $\textbf{x}^i$;\\
        \For{$j\leftarrow1$ \rm{\textbf{to}} $k$}{
            \uIf{$r = r_1$}{
                $d^i_j\leftarrow$ $-\sqrt{\sum_{l=1}^{m}(f_l(\textbf{x}^i)-(p^*)^j_l})^2$\;
            }
            \uElseIf{$r = r_2$}{
                $d^i_j\leftarrow$ $\sqrt{\sum_{l=1}^{m}\text{max}(f_l(\textbf{x}^i)-(p^*)^j_l}, 0)^2$\;
            }
            \Else{
                $d^i_j\leftarrow$ $\sqrt{\sum_{l=1}^{m}(f_l(\textbf{x}^i)-(p^*)^j_l})^2$\;
            }

        }
    }
    \textbf{Return} $d$.
\end{algorithm}

\subsection{Generating Offspring}
\label{section_3_4}
The process of generating offspring in the proposed algorithm is similar to that in genetic algorithms, in addition to the selection of individuals for filling up the gene pool from which the parent solutions are selected to generate offspring. In this subsection, the processes of generating offspring are elaborated in Steps~\ref{step_generate_offspting_1}-\ref{step_generate_offspting_5}. Then, the details for filling up the gene pool  are presented in Algorithm~\ref{alg_filling_up_gene_pool}.

\begin{enumerate}[Step 1:]
  \item Select solutions from the current population to fill up the gene pool, until it is full.
  \label{step_generate_offspting_1}
  \item Select two parent solutions from the gene pool and remove them from the gene pool.
  \label{step_generate_offspting_2}
  \item Employ the Simulated binary crossover (SBX) operator to generate offspring with the selected parent solutions.
  \item Employ the polynomial mutation operator to mutate the generated offspring.
  \label{step_generate_offspting_4}
  \item Repeat Steps~\ref{step_generate_offspting_2}-\ref{step_generate_offspting_4} until the gene pool is empty.
  \label{step_generate_offspting_5}
\end{enumerate}
\begin{algorithm}[htp]
  \caption{Filling Up the Gene Pool}
  \label{alg_filling_up_gene_pool}
    \KwIn{Current population $P_t$; Gene pool size $g$.}
    \KwOut{Gene pool $G$.}
    $G\leftarrow$ $\emptyset$;\\
    \While{the size of $G$ is less than $g$}{
        $\{\textbf{x}^1, \textbf{x}^2\} \leftarrow$ Randomly select two individuals from $P_t$;\\
        \label{alg_filling_up_gene_pool_line1}
        $r_{\textbf{x}^1} \leftarrow$ Obtain the rank of ${\textbf{x}^1}$;\\
        \label{alg_filling_up_gene_pool_line2}
        $r_{\textbf{x}^2} \leftarrow$ Obtain the rank of ${\textbf{x}^2}$;\\
        $\textbf{d}^1 \leftarrow$ Obtain the proximity distances of ${\textbf{x}^1}$;\\
        $\textbf{d}^2 \leftarrow$ Obtain the proximity distances of ${\textbf{x}^2}$;\\
        \label{alg_filling_up_gene_pool_line5}
        \uIf{$r_{\textbf{x}^1}<r_{\textbf{x}^2}$}{
        \label{alg_filling_up_gene_pool_line6}
            $G\leftarrow$ $G\cup\textbf{x}^1$;\\
        }\uElseIf{$r_{\textbf{x}^1}>r_{\textbf{x}^2}$}{
            $G\leftarrow$ $G\cup\textbf{x}^2$;\\
            \label{alg_filling_up_gene_pool_line8}
        }\Else{
            \uIf{$min(\textbf{d}^1)<min(\textbf{d}^2)$}{
            \label{alg_filling_up_gene_pool_line9}
                $G\leftarrow$ $G\cup\textbf{x}^1$;\\
            }\uElseIf{$min(\textbf{d}^1)>min(\textbf{d}^2)$}{
                $G\leftarrow$ $G\cup\textbf{x}^2$;\\
                \label{alg_filling_up_gene_pool_line12}
            }\Else{
                \label{alg_filling_up_gene_pool_line13}
                $\textbf{x} \leftarrow$ Randomly select one individual from $\{\textbf{x}^1, \textbf{x}^2\}$;\\
                $G\leftarrow$ $G\cup\textbf{x}$;\\
                \label{alg_filling_up_gene_pool_line15}
            }
        }
    }
    \textbf{Return} $G$.
\end{algorithm}

The binary tournament selection~\cite{miller1995genetic} approach is employed in Algorithm~\ref{alg_filling_up_gene_pool} to select individuals from the current population to fill up the gene pool. In other words, the binary tournament selection~\cite{miller1995genetic} approach is employed to select individuals from the current population. Specifically, two individuals which are denoted by $\textbf{x}^1$ and $\textbf{x}^2$ are randomly selected from the current population first (line~\ref{alg_filling_up_gene_pool_line1}). Then, their ranks and the proximity distances are obtained (lines~\ref{alg_filling_up_gene_pool_line2}-\ref{alg_filling_up_gene_pool_line5}). Next, the individual with smaller rank value is selected to be copied to the gene pool (lines~\ref{alg_filling_up_gene_pool_line6}-\ref{alg_filling_up_gene_pool_line8}). If $\textbf{x}^1$ and $\textbf{x}^2$ have the same rank values, the individual who has the smaller minimal proximity distance is selected (lines~\ref{alg_filling_up_gene_pool_line9}-\ref{alg_filling_up_gene_pool_line12}). Otherwise, an individual from $\textbf{x}^1$ and $\textbf{x}^2$ is randomly selected being as one potential parent solution to be put into the gene pool (lines~\ref{alg_filling_up_gene_pool_line13}-\ref{alg_filling_up_gene_pool_line15}). When the gene pool is full, two parent solutions are randomly selected from the gene pool for generating offspring, and then these selected parent solutions are removed from the gene pool until the gene pool is empty. Noted here that, the SBX~\cite{deb1994simulated} and the polynomial mutation~\cite{deb2001multi} operators are employed for the corresponding crossover and mutation operations in the proposed algorithm. It has been reported that two solutions selected in a large search space is not necessary to generate promising offspring~\cite{purshouse2007evolutionary,adra2011diversity}. Generally, two ways can be employed to solve this problem. One is the mating restriction method to limit the offspring to be generated by the neighbor solutions~\cite{deb1989investigation}. The other one is to use SBX with a large distribution index~\cite{deb2014evolutionary}. In the proposed algorithm, the latter one is utilized due to its simplicity.

\subsection{Environmental Selection}
\label{section_3_5}
When the offspring have been generated, the size of the current population is greater than that of the available slots. As a consequence, the environmental selection takes effects to select a set of representatives to survive to the next generation. In summary, the individuals are selected from the current population according to their assigned rank values and proximity distances. For convenience, it is assumed that there are $N$ available slots, the selected individuals are to be stored in $P_{t+1}$, and the individuals with ranks $r_1$, $r_2$, as well as $r_3$ are grouped into $F_{r_1}$, $F_{r_2}$, and $F_{r_3}$ non-dominated fronts, respectively. To be specific, the counter $i$ is increased by one until $\sum_{j=1}^i|F_{r_j}| > N$ where $|\cdot|$ is a countable operator. If $\sum_{j=1}^{i-1}|F_{r_j}| $ is equal to $N$, the individuals in $F_{r_1},\cdots, F_{r_{i-1}}$ are copied into $P_{t+1}$ and the environmental selection is terminated. Otherwise, the individuals in $F_{r_1},\cdots, F_{r_{i-1}}$ are copied into $P_{t+1}$ first, then $A=N-\sum_{j=1}^{i-1}|F_{r_j}|$ individuals are selected from $F_{r_i}$.
In summary, the details of the environmental selection are presented in Algorithm~\ref{alg_environment_selection}. Furthermore, line~\ref{alg_select_A_individuals} is confirmed by finding $A$ individuals who have the minimal total proximity distances to the $A$ reference points $r$ (line~\ref{alg_select_A_reference_points}), which involves a linear assignment problem (LAP). In the proposed algorithm, the Hungarian method~\cite{jonker1986improving} is employed to solve this LAP.

\begin{algorithm}
  \caption{Environmental selection}
  \label{alg_environment_selection}
    \KwIn{$F_{r_1}$, $F_{r_2}$, and $F_{r_3}$; Available slots size $N$.}
    \KwOut{$P_{t+1}$.}
    $P_{t+1}\leftarrow$ $\emptyset$;\\
    $i\leftarrow$ $1$;\\
    \While{$|P_{t+1}| + |F_{r_i}| < N$}
    {
        $P_{t+1}\leftarrow$ $P_{t+1} \cup F_{r_i}$;\\
        $i\leftarrow$ $i+1$;\\
    }
    \uIf{$|P_{t+1}| + |F_{r_i}| = N$}{
        $P_{t+1}\leftarrow$ $P_{t+1} \cup F_{r_i}$;\\
    }\Else{
        $r\leftarrow$ Uniformly select $A=N-\sum_{j=1}^{i-1}|F_{r_j}|$ reference points from $p^*$;\\
        \label{alg_select_A_reference_points}
        $R\leftarrow$ Select $A$ individuals from $F_{r_i}$;\\
        \label{alg_select_A_individuals}
        $P_{t+1}\leftarrow$ $P_{t+1}\cup R$;
    }

    \textbf{Return} $P_{t+1}$.
\end{algorithm}

\subsection{Computational Complexity}
\label{section_3_6}
In this subsection, the computational complexity of the proposed algorithm is analyzed. For convenience, it is assumed that the problem to be optimized is with $m$ objectives, $n$ decision variables, $N$ desired solutions for decision-makers, and the computational complexity is analyzed in the context of Algorithm~\ref{alg_the_proposed_algorithm}. To estimate each extreme point, the genetic algorithm is employed, and the SBX as well as polynomial mutation are used as the genetic operators. Furthermore, it is assumed that the population size for estimating extreme points is set to be $N$, and the generation is set to be $t_1$. Consequently, the total computation cost of uniformly generating reference points for IGD indicator (line~\ref{alg_framework_line_1}) is $O(t_1m^2N)$. Furthermore, lines~\ref{alg_framework_line_2} and~\ref{alg_framework_line_3} require $O(nN)$ and $O(mN)$ computations, respectively. Because the number of the reference points is equal to that of the desired solutions, the computational complexity of assigning ranks and proximity distances in line~\ref{alg_framework_line_4} are $O(mN^2)$ and $O(nN^2)$, respectively. Furthermore, generating offspring (line~\ref{alg_framework_line_5}) needs $O(\frac{N}{2}(n+n))$ computations because the size of gene pool is set to be $N$. Since only the fitness, ranks and the proximity distances of the generated offspring need to be calculated, as a consequence, lines~\ref{alg_framework_line_6} and~\ref{alg_framework_line_7} consume $O(\frac{N}{2}m)$ and $O(\frac{N}{2}Nm)$ + $O(\frac{N}{2}Nn)$, respectively. In the environmental selection, the best case scenario in computational complexity is $O(N)$, while the worst is $O(N^3)$ given that $N$ individuals are linearly assignment to the reference points. Furthermore, it is considered common that $N$ is greater than $n$, and $N>>m$ in MaOPs. Therefore, lines~\ref{alg_framework_start_while}-\ref{alg_framework_end_while} overall need $O(tN^3)$ computations with the generation $t$. In summary, the computational complexity of the proposed algorithm is $O(tN^3)$ where $t$ is the number of the generation and $N$ is the number of solutions.

\subsection{Discussions}
\label{section_3_7}
Loss of selection pressure is a major issue for traditional MOEAs in effectively solving MaOPs because of the traditional domination comparisons between individuals giving a large proportion of non-dominated solutions. In the proposed algorithm, the dominance relation of all the individuals are compared to the reference points which are employed for the calculation of IGD indicator. However, the exact reference points which are uniformly distributed in the PF are difficult to obtain. For this purpose, a set of points which are evenly distributed in the Utopian PF are sampled. Furthermore, in order to address this inefficiency given by these approximated reference points, three proximity distances are designed according to their dominance relation to the approximated reference points. This is in hope that less value of the proximity distance means that the corresponding individual is with a better proximity. Specifically, if the solutions with rank $r_2$ are still with the distance calculation of that with ranks $r_1$ or $r_3$, the convergence will be lost in the proposed algorithm~\cite{ishibuchi2015modified}.

When the number of solutions to be selected is larger than the available slots, the representatives are chosen from a global view in the proposed algorithm. For convenience of understanding, it is first assumed that $a$ representatives need to be selected from $b$ solutions where $b>a$. Then the selection of $a$ representatives is simultaneously considered by the calculation of IGD indicator, as oppose to choosing one by one. Simultaneously selecting $a$ representatives involves a linear assignment problem. By this linear assignment, each selected reference point can have one distinct individual, which improves the diversity and the convergence simultaneously and this conclusion can also be found in literatures~\cite{berenguer2015evolutionary,lopez2016igd+,rodriguez2012new}. If the individuals are selected by finding the individual who has the least distance to the reference points, the diversity is not necessarily guaranteed.

\section{Experiments}
\label{section_4}
To evaluate the performance of the proposed algorithm in solving MaOPs, a series of experiments is performed. Particularly, NSGA-III~\cite{deb2014evolutionary}, MOEA/D~\cite{zhang2007moea}, HypE~\cite{bader2011hype}, RVEA~\cite{cheng2016reference}, and KnEA~\cite{zhang2015knee} are selected as the state-of-the-art peer competitors. Although the IGD$^{+}$-EMOA can be viewed as the peer algorithm based on IGD indicator, it is merely capable of solving MaOPs with no more than $8$ objectives. As a consequence, IGD$^{+}$-EMOA is excluded from the list of peer competitors in our experiments.

The remaining of this section is organized as follows. At first, the selected benchmark problems used in this experiment are introduced. Then, the chosen performance metric is given to measure the quality of the approximate Pareto-optimal solutions generated by the competing algorithms. Next, the parameter settings employed in all the compared algorithms are listed, and experimental results measured by the considered performance metric are presented and analyzed. Finally, the performance of the proposed algorithm in solving a real-world MaOP is shown (in Section III of the Supplemental Materials), and the performance on the proposed DNPE in estimating nadir point is empirically investigated.

\subsection{Benchmark Test Problems}
The widely used scalable test problems DTLZ1-DTLZ7 from the DTLZ benchmark test suite~\cite{deb2005scalable} and WFG1-WFG9 from the WFG benchmark test suite~\cite{huband2006review} are employed in our experiments. Specifically, each objective function in one given $m$-objective test problem of DTLZ has $n=k+m-1$ decision variables, and $k$ is set to be $5$ for DTLZ1, $10$ for DTLZ2-DTLZ6, and $20$ for DTLZ7 problems. Moreover, each objective function of a given problem in WFG test suite has $n=k+l$ decision variables, and $k$ is set to be $(m-1)$ and $l$ is set to be $20$ based on the suggestion from~\cite{huband2006review}.
\subsection{Performance Metric}
The widely used Hypervolume (HV)~\cite{zitzler1999multiobjective} which simultaneously measures the convergence and diversity of the MaOEAs is selected as the performance metric in these experiments. Specifically, the reference points for the calculation of HV are set to be $\{1,\cdots,1\}$ for DTLZ1, $\{2,\cdots,2\}$ for DTLZ2-DTLZ6, and $\{3,\cdots,2m+1\}$ for DTLZ7 as well as WFG1-WFG9 test problems. Please note that the solutions are discarded for the calculation of HV when they are dominated by the predefined reference points. Because the computational cost increases significantly as the number of objectives grows, Monte Carlo simulation~\cite{bader2011hype}\footnote{The source code is available at:~\url{http://www.tik.ee.ethz.ch/sop/download/supplementary/hype/}.} is applied for the calculation when $m\geq 10$, otherwise the exact approach proposed in~\cite{while2012fast} is utilized\footnote{The source code is available at:~\url{http://www.wfg.csse.uwa.edu.au/hypervolume/}.}. In our experiments, all the HV values are normalized to $[0,1]$ by dividing the HV value of the origin with the corresponding reference point. Moreover, higher HV values indicate a better performance of the corresponding MaOEA.

\subsection{Parameter Settings}
In this subsection, the baseline parameter settings which are adopted by all the compared MaOEAs are declared first. Then the special parameter settings required by each MaOEA are provided.
\subsubsection{Number of Objectives} Test problems with $8$, $15$, and $20$ objectives are considered in the experiments because the proposed algorithm aims specifically at effectively solving MaOPs.
\subsubsection{Number of Function Evaluations} and Stop Criterion All compared algorithms are individually executed $30$ independent times. The maximum number of function evaluations for each compared MaOEA in one independent run is set to be $2.3\times 10^6$, $4.3\times 10^6$, and $5.5\times 10^6$ for $8$-, $15$-, and $20$-objective, respectively, which is employed as the termination criterion. Noted that, the parameter settings here are set based on the convention that the maximum generations for the MaOEAs with more than $10$ objectives are generally in the order of $10^3$ (the generation number set here is approximately to $1,200$). Because of the proposed algorithm includes the phases of nadir point estimation and the optimization for MaOPs, the function evaluations specified here will be shared by these two phases for a fair comparison.
\subsubsection{Statistical Approach} Because of the heuristic characteristic of the peer evolutionary algorithms, all the results, which are measured by the performance metric over $30$ independent runs for each competing algorithm, are statistically evaluated. In this experiment, the Mann-Whitney-Wilcoxon rank-sum test~\cite{steel1997principles} with a $5\%$ significance level is employed for this purpose.
\begin{table}[ht]
\caption{configurations of two layers setting.}
\label{two_layers_setting}
\begin{center}
\begin{tabular}{p{0.05\columnwidth}<{\centering}|p{0.3\columnwidth}<{\centering}|p{0.4\columnwidth}<{\centering}}
\hline
$m$ & \#  of divisions & \#  of population \\
\hline
8 & 3,3 & 240\\
15 & 2,2 & 240\\
20 & 2,1 & 230\\
\hline
\end{tabular}
\end{center}
\end{table}
\subsubsection{Population Size} In principle, the population size can be arbitrarily assigned. However, the population size of the proposed algorithm, NSGA-III, MOEA/D, and RVEA depends on the number of the associated reference points or reference vectors. For a fair comparison, the sizes of population in HypE and KnEA are set to be the same as that in others. In the experiment, the reference points and reference vectors are sampled with the two-layer method~\cite{deb2014evolutionary} and the configurations are listed in Table~\ref{two_layers_setting}.

\begin{table*}[!htb]
\caption{HV results of MaOEA/IGD against NSGA-III, MOEA/D, HypE, RVEA, and KnEA over DTLZ1-DTLZ7 with $8$-, $15$-, and $20$-objective.}
\label{hv_results_on_dtlz1-4}
\begin{center}
\begin{tabular}{c|c|c|c|c|c|c|c|c}
\hline
& & &MaOEA/IGD&NSGA-III&MOEA/D&HypE&RVEA&KnEA\\
\hline
\multirow{6}{*}{\rotatebox{90}{\textbf{Linear}}}&\multirow{3}{*}{DTLZ1}
&8&\textbf{0.9998(2.93E-4)}&0.9964(6.12E-4)(+)&0.9996(3.52E-5)(+)&0.7213(4.31E-1)(+)&0.9992(2.15E-4)(+)&0.9921(4.21E-5)(+)\\
\cline{3-9}
& &15&0.9990(3.18E-3)&0.9984(7.23E-4)(+)&0.9987(3.20E-4)(+)&0.6922(5.45E-1)(+)&0.9992(3.99E-3)(-)&\textbf{0.9994(2.91E-4)(-)}\\
\cline{3-9}
& &20&\textbf{0.9990(2.32E-3)}&0.9983(3.82E-4)(+)&0.9977(7.24E-4)(+)&0.7672(3.88E-1)(+)&0.9989(3.68E-4)(=)&0.9890(3.80E-4)(+)\\
\cline{2-9}
&\multirow{3}{*}{DTLZ7}
&8&\textbf{0.6999(8.97E-3)}&0.6959(2.72E-5)(=)&0.5439(6.37E-5)(+)&0.2122(4.82E-2)(+)&0.6894(4.24E-4)(=)&0.5466(8.12E-2)(+)\\
\cline{3-9}
& &15&0.3592(1.20E-2)&0.2769(2.33E-5)(+)&0.2119(2.42E-2)(+)&0.1999(7.68E-5)(+)&\textbf{0.4070(9.50E-4)(-)}&0.2804(2.11E-2)(+)\\
\cline{3-9}
& &20&\textbf{0.5261(7.46E-4)}&0.2348(3.29E-4)(+)&0.4325(1.12E-4)(+)&0.0986(5.31E-4)(+)&0.5206(7.58E-2)(+)&0.5166(8.24E-5)(+)\\
\hline
\multirow{15}{*}{\rotatebox{90}{\textbf{Concave}}}&\multirow{3}{*}{DTLZ2}
&8&0.7174(3.96E-3)&\textbf{0.8132(2.78E-3)(-)}&0.5221(3.83E-3)(+)&0.1121(3.34E-2)(+)&0.6821(3.68E-3)(+)&0.7320(3.66E-3)(+)\\
\cline{3-9}
& &15&\textbf{0.9268(2.62E-3)}&0.8832(9.11E-3)(+)&0.3329(1.73E-2)(+)&0.0892(4.12E-2)(+)&0.9020(3.48E-3)(+)&0.8599(6.82E-3)(+)\\
\cline{3-9}
& &20&0.8905(6.80E-3)&\textbf{0.9660(3.23E-3)(-)}&0.3298(2.10E-2)(+)&0.0633(5.32E-2)(+)&0.9443(3.19E-3)(-)&0.9307(4.38E-3)(-)\\
\cline{2-9}
&\multirow{3}{*}{DTLZ3}
&8&0.4664(9.25E-2)&0.0055(3.80E-4)(+)&\textbf{0.5169(5.68E-3)(-)}&0.0085(0.76E-5)(+)&0.4572(0.54E-3)(=)&0.3537(5.31E-4)(+)\\
\cline{3-9}
& &15&0.6984(6.68E-2)&0.0091(0.78E-5)(+)&0.3030(4.43E-3)(+)&0.0133(1.07E-5)(+)&\textbf{0.7183(9.62E-2)(-)}&0.5961(0.05E-2)(+)\\
\cline{3-9}
& &20&\textbf{0.7476(7.52E-2)}&0.0002(6.48E-4)(+)&0.2162(4.51E-4)(+)&0.0065(5.47E-4)(+)&0.6491(2.96E-2)(+)&0.7317(7.45E-4)(+)\\
\cline{2-9}
&\multirow{3}{*}{DTLZ4}
&8&\textbf{0.8338(3.31E-3)}&0.8187(6.22E-4)(+)&0.5322(5.87E-2)(+)&0.2537(2.08E-4)(+)&0.8159(3.01E-4)(+)&0.8302(4.71E-3)(=)\\
\cline{3-9}
& &15&\textbf{0.9548(1.66E-3)}&0.9537(4.24E-4)(=)&0.3150(5.08E-3)(+)&0.1957(0.86E-4)(+)&0.9267(2.62E-5)(+)&0.9188(8.01E-3)(+)\\
\cline{3-9}
& &20&0.9824(1.33E-3)&\textbf{0.9947(1.37E-3)(-)}&0.2755(7.21E-5)(+)&0.2101(1.07E-4)(+)&0.9854(6.54E-2)(-)&0.9797(4.94E-3)(+)\\
\cline{2-9}
&\multirow{3}{*}{DTLZ5}
&8&0.4190(0.64E-3)&0.3908(7.67E-3)(+)&0.3174(7.15E-5)(+)&0.0451(2.24E-5)(+)&0.3474(0.88E-3)(+)&\textbf{0.6401(2.65E-4)(-)}\\
\cline{3-9}
& &15&\textbf{0.2677(9.71E-3)}&0.2178(5.34E-5)(+)&0.1821(8.85E-2)(+)&0.0418(8.99E-5)(+)&0.1379(7.84E-4)(+)&0.1257(3.46E-3)(+)\\
\cline{3-9}
& &20&0.2101(5.57E-3)&0.3390(0.44E-2)(-)&0.1790(0.99E-4)(+)&0.0423(4.90E-2)(+)&0.3606(5.52E-2)(-)&\textbf{0.4139(5.49E-2)(-)}\\
\cline{2-9}
&\multirow{3}{*}{DTLZ6}
&8&\textbf{0.7202(8.94E-4)}&0.2866(0.22E-5)(+)&0.3037(7.95E-3)(+)&0.0548(7.36E-4)(+)&0.2467(9.72E-2)(+)&0.4547(4.17E-3)(+)\\
\cline{3-9}
& &15&\textbf{0.7756(8.36E-5)}&0.4385(7.14E-3)(+)&0.6748(4.56E-2)(+)&0.1957(0.86E-4)(+)&0.4918(9.97E-2)(+)&0.7314(1.56E-2)(+)\\
\cline{3-9}
& &20&0.8639(0.20E-3)&\textbf{0.9081(8.32E-2)(-)}&0.5170(5.20E-4)(+)&0.1080(6.17E-2)(+)&0.4438(2.39E-4)(+)&0.3002(5.79E-5)(+)\\
\hline
\multicolumn{4}{c|}{+/=/-}&14/2/5&20/0/1&21/0/0&12/3/6&16/1/4\\
\hline
\end{tabular}
\end{center}
\end{table*}
\subsubsection{Genetic Operators}
The SBX~\cite{deb1994simulated} and polynomial mutation~\cite{deb1996combined} are employed as the genetic operators. Moreover, the probabilities of the crossover and mutation are set to be $1$ and $1/n$, respectively. The distribution indexes of mutation and crossover are set to be $20$, in addition to NSGA-III whose mutation distribution index is specifically set to be $30$ based on the recommendation from its developers~\cite{deb2014evolutionary}.

In solving the proposed nadir point estimation method for constructing the Utopian PF, evolutionary algorithm is employed. To be specific, the SBX and polynomial mutation, both of whose distribution index are set to be $20$, and probabilities for crossover and mutation are set to be $0.9$ and $1/n$, respectively, are utilized as the genetic operators. In addition, the population sizes are set to be the same to those in Table~\ref{two_layers_setting}, and the numbers of generations for all are set to be $1,000$. Besides, the balance parameter $\lambda$ in~(\ref{equ_extrme_point}) is specified as $100$.

\subsection{Experimental Results and Analysis}
In this subsection, the results, which are generated by competing algorithms over considered test problems with specific objective numbers and then measured by the selected performance metric, are presented and analyzed to highlight the superiority of the proposed algorithm in addressing MaOPs. Specifically, the mean values as well as the standard deviations of HV results over DTLZ1-DTLZ7 and WFG1-WFG9 test problems are listed in Tables~\ref{hv_results_on_dtlz1-4} and~\ref{hv_results_on_wfg1-9}, respectively. Furthermore, the numbers with bold face imply the best mean values over the corresponding test problem with a given objective number (the second and third columns in Tables~\ref{hv_results_on_dtlz1-4} and \ref{hv_results_on_wfg1-9}) against all compared algorithms. Moreover, the symbols ``+,'' ``-,'' and ``='' indicate whether the null hypothesis of the results, which are generated by the proposed algorithm and corresponding compared peer competitor, is accepted or rejected with the significance level $5\%$ by the considered rank-sum test. In addition, the last rows in Tables~\ref{hv_results_on_dtlz1-4} and \ref{hv_results_on_wfg1-9} present the summarizations indicating how many times the proposed algorithm performs better than, worse than or equal to the chosen peer competitor, respectively. In order to conveniently investigate the experimental results of the well-designed proximity distance assignments in the proposed MaOEA/IGD and the conclusion in Section~\ref{section_3_3}, test problems are group into ``Convex,'' ``Linear,'' and ``Concave'' based on the respective test problem features, and displayed in the first columns of Tables~\ref{hv_results_on_dtlz1-4} and \ref{hv_results_on_wfg1-9}. Noted that, although the PFs of DTLZ7 and WFG1 are mixed, they are classified into the ``Linear'' category due to their PF shapes being more similar to linear.

From the results measured by HV on DTLZ1-DTLZ7 test problems (Table~\ref{hv_results_on_dtlz1-4}), it is clearly shown that MaOEA/IGD achieves the best performance among its peer competitors upon $8$- and $20$-objective DTLZ1 and DTLZ7, while performs slightly worse upon $15$-objective DTLZ1 by KnEA and DTLZ7 by RVEA. Furthermore, MaOEA/IGD also wins the best scores on $8$- and $15$-objective DTLZ4 and DTLZ6, but is defeated by NSGA-III upon these two problems with $20$-objective. Although NSGA-III and KnEA show better performance upon $8$- and $20$-objective DTLZ2 and DTLZ5, MaOEA/IGD is the winner upon $15$-objective DTLZ2 and DTLZ5. In addition, MaOEA/IGD achieves the best score upon $20$-objective DTLZ3.

\begin{table*}[!htb]
\caption{HV results of MaOEA/IGD against NSGA-III, MOEA/D, HypE, RVEA, and KnEA over WFG1-WFG9 with $8$-, $15$-, and $20$-objective.}
\label{hv_results_on_wfg1-9}
\begin{center}
\begin{tabular}{c|c|c|c|c|c|c|c|c}
\hline
& & &MaOEA/IGD&NSGA-III&MOEA/D&HypE&RVEA&KnEA\\
\hline
\multirow{3}{*}{\rotatebox{90}{\textbf{Convex}}}&\multirow{3}{*}{WFG2}
&8&\textbf{0.9839(2.00E-2)}&0.9587(9.10E-3)(+)&0.9474(9.09E-2)(+)&0.9514(5.92E-4)(+)&0.9380(3.33E-3)(+)&0.9686(8.53E-2)(+)\\
\cline{3-9}
& &15&0.9362(7.76E-3)&\textbf{0.9672(3.16E-2)(-)}&0.9402(7.00E-4)(-)&0.6216(6.25E-3)(+)&0.9475(9.71E-4)(-)&0.9360(8.37E-3)(=)\\
\cline{3-9}
& &20&\textbf{0.9782(2.52E-2)}&0.9624(0.11E-2)(+)&0.9460(5.73E-2)(+)&0.8354(7.90E-4)(+)&0.9657(4.04E-2)(+)&0.8106(5.11E-2)(+)\\
\hline
\multirow{6}{*}{\rotatebox{90}{\textbf{Linear}}}&\multirow{3}{*}{WFG1}
&8&\textbf{0.9578(1.30E-1)}&0.9255(9.48E-2)(+)&0.9454(0.61E-4)(+)&0.6528(5.85E-4)(+)&0.8383(2.85E-4)(+)&0.5847(8.28E-2)(+)\\
\cline{3-9}
& &15&0.9354(5.02E-3)&0.9536(3.29E-3)(-)&0.9405(6.50E-2)(=)&0.6395(9.75E-3)(+)&\textbf{0.9538(1.84E-3)(-)}&0.9373(4.98E-2)(+)\\
\cline{3-9}
& &20&\textbf{0.9806(2.64E-2)}&0.9405(8.01E-2)(+)&0.9593(1.43E-2)(+)&0.6340(4.78E-2)(+)&0.9020(5.43E-2)(+)&0.9374(8.84E-2)(+)\\
\cline{2-9}
& \multirow{3}{*}{WFG3}
&8&0.8615(1.08E-1)&0.8423(7.55E-2)(+)&0.8521(7.42E-4)(=)&0.5660(8.31E-3)(+)&0.8457(2.34E-2)(+)&\textbf{0.8618(1.57E-2)(=)}\\
\cline{3-9}
& &15&0.3346(4.58E-3)&0.5091(4.10E-2)(-)&0.3393(1.32E-2)(=)&0.2852(5.41E-3)(+)&\textbf{0.5188(2.43E-2)(-)}&0.5076(8.26E-4)(-)\\
\cline{3-9}
& &20&\textbf{0.5750(3.13E-2)}&0.5313(3.89E-2)(+)&0.4863(4.29E-2)(+)&0.2918(9.56E-3)(+)&0.3425(5.73E-2)(+)&0.4579(8.50E-2)(+)\\
\hline
\multirow{18}{*}{\rotatebox{90}{\textbf{Concave}}}&\multirow{3}{*}{WFG4}
&8&0.7800(2.64E-2)&\textbf{0.7877(7.02E-2)(=)}&0.7507(3.75E-4)(+)&0.7229(9.74E-3)(+)&0.7648(7.29E-2)(+)&0.7715(1.74E-4)(+)\\
\cline{3-9}
& &15&0.8314(5.76E-3)&0.6315(0.01E-2)(+)&0.8414(0.03E-2)(-)&0.4801(0.87E-2)(+)&0.8154(2.61E-3)(+)&\textbf{0.8690(0.23E-2)(-)}\\
\cline{3-9}
& &20&\textbf{0.8108(3.86E-2)}&0.7885(4.68E-3)(+)&0.7707(8.61E-2)(+)&0.7309(4.67E-2)(+)&0.8094(8.35E-3)(=)&0.4794(7.43E-2)(+)\\
\cline{2-9}
& \multirow{3}{*}{WFG5}
&8&0.8653(1.10E-1)&0.7993(4.57E-4)(+)&0.4429(6.68E-2(+)&0.4915(6.99E-3)(+)&0.8638(6.51E-3)(+)&\textbf{0.8741(3.09E-2)(-)}\\
\cline{3-9}
& &15&\textbf{0.8335(6.31E-3)}&0.6408(1.69E-2)(+)&0.3397(0.01E-2)(+)&0.4351(4.18E-2)(+)&0.7553(4.88E-2)(+)&0.8276(1.60E-3)(+)\\
\cline{3-9}
& &20&\textbf{0.8905(6.28E-3)}&0.8022(9.87E-4)(+)&0.4964(0.84E-2)(+)&0.3125(2.50E-2)(+)&0.7946(9.13E-4)(+)&0.7258(6.64E-3)(+)\\
\cline{2-9}
& \multirow{3}{*}{WFG6}
&8&0.9785(3.12E-2)&0.9918(8.77E-2)(-)&0.9488(8.06E-4(+)&0.9261(4.61E-2)(+)&\textbf{0.9976(6.97E-4)(-)}&0.9876(3.68E-2)(-)\\
\cline{3-9}
& &15&\textbf{0.9357(8.22E-3)}&0.8756(5.13E-3)(+)&0.8327(2.41E-4)(+)&0.8406(2.60E-3)(+)&0.8538(0.21E-2)(+)&0.9223(8.21E-2)(+)\\
\cline{3-9}
& &20&\textbf{0.8854(1.82E-2)}&0.8189(2.62E-2)(+)&0.8489(5.80E-2)(+)&0.7633(8.78E-4)(+)&0.7968(5.83E-2)(+)&0.7502(5.00E-3)(+)\\
\cline{2-9}
& \multirow{3}{*}{WFG7}
&8&\textbf{0.8858(5.43E-3)}&0.8118(7.25E-2)(+)&0.7430(8.58E-4)(+)&0.7416(3.48E-2)(+)&0.8192(2.51E-2)(+)&0.7635(5.82E-2)(+)\\
\cline{3-9}
& &15&0.8352(6.48E-3)&\textbf{0.8780(3.82E-2)(-)}&0.7343(7.92E-4)(+)&0.4030(8.39E-3)(+)&0.6366(1.79E-4)(+)&0.5463(1.70E-3)(+)\\
\cline{3-9}
& &20&0.7919(3.89E-3)&0.8482(1.35E-4)(-)&0.4844(9.14E-3)(+)&0.6418(6.41E-3)(+)&\textbf{0.8706(5.71E-1)(-)}&0.8116(9.03E-3)(-)\\
\cline{2-9}
& \multirow{3}{*}{WFG8}
&8&0.6839(1.78E-2)&\textbf{0.6869(1.39E-2)(-)}&0.4405(3.49E-3)(+)&0.3155(1.51E-3)(+)&0.5908(5.04E-3)(+)&0.6850(5.72E-3)(-)\\
\cline{3-9}
& &15&\textbf{0.7340(7.08E-3)}&0.5470(5.14E-3)(+)&0.3412(8.14E-4)(+)&0.2065(0.97E-2)(+)&0.6455(5.90E-4)(+)&0.6246(1.24E-2)(+)\\
\cline{3-9}
& &20&0.7831(2.08E-2)&0.6855(5.75E-3)(+)&0.4928(9.16E-2)(+)&0.1117(4.95E-4)(+)&\textbf{0.7844(8.87E-3)(=)}&0.6027(4.21E-2)(+)\\
\cline{2-9}
& \multirow{3}{*}{WFG9}
&8&\textbf{0.7694(8.69E-0)}&0.7328(8.73E-3)(+)&0.4488(0.55E-3)(+)&0.3030(5.00E-2)(+)&0.7444(3.41E-2)(+)&0.7528(4.91E-2)(+)\\
\cline{3-9}
& &15&\textbf{0.8329(8.16E-3)}&0.6105(0.13E-2)(+)&0.3359(7.18E-4)(+)&0.2176(3.91E-2)(+)&0.7294(0.34E-3)(+)&0.6595(4.06E-2)(+)\\
\cline{3-9}
& &20&\textbf{0.7829(2.79E-2)}&0.7299(5.76E-4)(+)&0.4881(8.07E-4)(+)&0.1923(6.55E-1)(+)&0.7128(8.78E-3)(+)&0.7824(5.36E-2)(=)\\
\hline
\multicolumn{4}{c|}{+/=/-}&19/1/7&22/3/2&27/0/0&20/2/5&18/3/6\\
\hline
\end{tabular}
\end{center}
\end{table*}

The HV results from WFG1-WFG9 test problems generated by competing algorithms are listed in Table~\ref{hv_results_on_wfg1-9}. For $8$-objective WFG test problems, MaOEA/IGD shows a better performance on WFG1, WFG2, WFG7, and WFG9 than its peer competitors, and performs a little worse than that of KnEA on WFG5, RVEA on WFG6, and NSGA-III on WFG8 test problems. Although MaOEA/IGD does not show the best scores on WFG3 and WFG4, it obtains similar statistical results compared to the respective winners (i.e., KnEA and NSGA-III). For $15$-objective test problems, MaOEA/IGD shows a better performance on WFG5, WFG6, WFG8, and WFG9 than competing algorithms, while worse than RVEA on WFG2 and WFG3, NSGA-III on WFG2, and KnEA on WFG4. Although NSGA-III performs better than MaOEA/IGD on WFG7, MaOEA/IGD performs better than all other peer competitors. In addition, MaOEA/IGD wins over NSGA-III, MOEA/D, HypE, RVEA, and KnEA on $20$-objective WFG1, WFG2, WFG3, WFG4, WFG5, WFG6, and WFG9, but underperforms on WFG7 and WFG8 in which RVEA performs better.

Briefly, MaOEA/IGD wins 9 times out of the 12 comparisons upon the test problems whose PF shapes are linear (i.e., DTLZ1, DTLZ7, WFG1, and WFG3), which can be interpreted that the sampled reference points from the Utopian PF for the proposed algorithm are the Pareto-optimal solutions due to the linear feature of the PF, and the proximity distance assignment for the solutions with rank value $r_2$ has taken effects. Furthermore, MaOEA/IGD shows competitive performance on WFG2 test problem whose feature of the PF is convex. Because the sampled reference points on the Utopian PF are all non-dominated by the Pareto-optimal solutions, the proximity distances for solutions with rank $r_1$ in MaOEA/IGD take effects in this situation. In addition, it is no strange that MaOEA/IGD obtains better results on most of other test problems whose PF features are concave because the reference points utilized to maintain the diversity and convergence of the proposed algorithm dominate the solutions uniformly generated from the PF. In summary, the proposed algorithm shows considerable competitiveness against considered competing algorithms in addressing selected MaOPs with the results measured by the HV performance metric.

Theoretically, the major shortcoming of HV indicator against IGD is its much higher computational complexity. However, noted that from Tables~\ref{hv_results_on_dtlz1-4} and~\ref{hv_results_on_wfg1-9}, the proposed algorithm, which is designed based on the IGD indicator, outperforms HypE, which is motivated by the HV indicator, upon all test problems with the selected numbers of objectives, although the numbers of function evaluations regarding HypE is set to be a much large number. The deficiencies of HypE in this regard are explained as follows. First, it has been reported in~\cite{auger2009theory,ishibuchi2010many,yuan2016balancing,yuan2016new} that the HV result is largely affected by the nadir points of the problem to be optimized. In HypE, the nadir points are determined as the evolution continues. In this way, the obtained nadir point would be inaccurate during the early evolution process (the reasons have been discussed in reviewing the nadir point estimation approaches in Section~\ref{section_2}), which leads to the worse performance of HypE. Secondly, the HV results of HypE in solving MaOPs are estimated by Monte Carlo simulation, while the number of reference points in Monte Carlo simulation is critical to the successful performance~\cite{bader2011hype}. In practice, that number is unknown and unavailable of such may lead to a poor performance.

\subsection{Investigation on Nadir Point Estimation}
\label{section_experiment_nadir}

\begin{figure}[htp]
\begin{center}
\subfloat[DTLZ1]{\includegraphics[width=0.75\columnwidth]{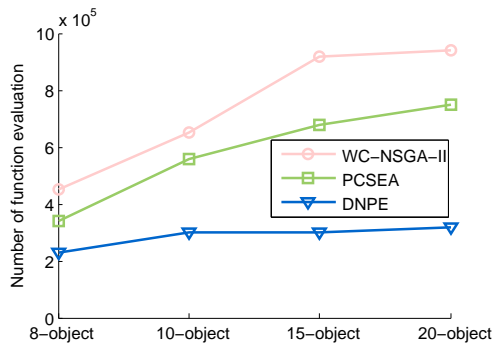}%
\label{fig_nadir_dtlz1}}
\hfil
\subfloat[DTLZ2]{\includegraphics[width=0.75\columnwidth]{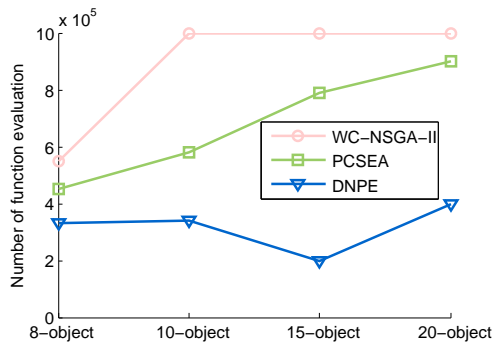}%
\label{fig_nadir_dtlz2}}
\hfil
\subfloat[WFG2]{\includegraphics[width=0.75\columnwidth]{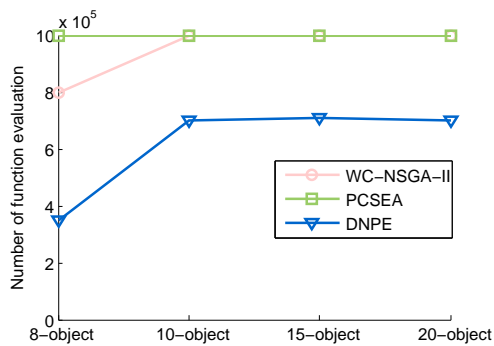}%
\label{fig_nadir_wfg2}}

\caption{The numbers of function evaluations performed by WC-NSGA-II, PCSEA, and DNPE on DTLZ1, DTLZ2, and WFG2 with $8$-, $10$-, $15$-, and $20$-objective.}
\label{fig_nadir_point_comparison}
\end{center}
\end{figure}

In this subsection, we will investigate the performance of the proposed DNPE on estimating the nadir point. To be specific, two peer competitors including WC-NSGA-II and PCSEA which have been discussed in Section~\ref{section_2} are utilized to perform comparisons on selected test problems. In these comparisons, the numbers of function evaluations regarding each compared algorithm are counted until 1) the metric $E \leq 0.01$ formulated by~(\ref{equ_nadir_metric})
\begin{equation}
\label{equ_nadir_metric}
E=\sqrt{\sum_{i=1}^m(z^{nad}_i-z_i)^2/(z^{nad}_i-z^*_i)^2}
\end{equation}
where $z_i$ denotes the $i$-th element of the estimated nadir point derived from the extreme points generated by the compared algorithm or 2) the maximum function evaluation numbers $100,000$ is met. The experimental results for DTLZ1, DTLZ2, and WFG2 with $8$-, $10$-, $15$-, and $20$-objective are plotted in Fig.~\ref{fig_nadir_point_comparison}. Please note that the reason of choosing these three test problems is that they cover the various shapes of PF (i.e., DTLZ1, DTLZ2, and WFG2 are with linear, concave, and convex PF, respectively) and characteristics of objective value scales (i.e., DTLZ1 and DTLZ2 are with the same objective value scales while WFG2 is not). Specifically, the ideal points of DTLZ1, DTLZ2, and WFG2 are $\{0,\cdots,0\}$, and the nadir points are $\{0.5,\cdots, 0.5\}$, $\{1,\cdots,1\}$, and $\{2, 4, \cdots, 2m\}$, respectively. In addition, the population size is specified as $200$, the probabilities of SBX and polynomial mutation are set to be $0.9$ and $1/n$, and both of distribution index are set to be $20$. Because the proposed DNPE is based on the decomposition to estimate the nadir point, $E \leq 0.01/m$ and maximum function evaluation number with $100,000/m$ are set to be the stopping criteria for estimating each extreme point.

The results performed by compared nadir point estimation methods on $8$-, $10$-, $15$-, and $20$-objective DTLZ1, DTLZ2, and WFG2 are illustrated in Figs.~\ref{fig_nadir_dtlz1},~\ref{fig_nadir_dtlz2}, and~\ref{fig_nadir_wfg2}, respectively. It is clearly shown in Fig.~\ref{fig_nadir_dtlz1} that these compared algorithms find the satisfactory nadir points of the DTLZ1 which is with the linear PF within the predefined maximum function evaluation numbers, and the proposed DNPE takes the least numbers of function evaluations over the four considered objective numbers. Moreover, WC-NSGA-II cannot find the nadir point over DTLZ2 with concave PF and WFG2 with convex PF with $10$-, $15$-, and $20$-objective, and PCSEA cannot find the nadir point over WFG2 with different objective value scales, while the proposed DNPE performs well on both test problems with all considered objective numbers. In addition, the proposed DNPE is scalable to the objective number in the estimating nadir points of the MaOPs, which can be seen from Figs.~\ref{fig_nadir_dtlz1} and~\ref{fig_nadir_dtlz2}. In summary, the proposed DNPE shows quality performance in estimating nadir point of MaOPs with different PF features and objective scales.

\section{Conclusion and Future Works}
\label{section_5}
In this paper, an IGD indicator-based evolutionary algorithm is proposed for solving many-objective optimization problems. In order to obtain a set of uniformly distributed reference points for the calculation of the IGD indicator, a decomposition-based nadir point estimation method is designed to construct the Utopian PF in which the reference points can be easily sampled. For solving the deficiency of the Utopian PF being as the PF in the phase of sampling the reference points, one rank assignment mechanism is proposed to compare the dominance relation of the solutions to the reference points, based on which three types of proximity distance assignments are designed to distinct the quality of the solutions with the same front rank values. In addition, the linear assignment principle is utilized as the selection mechanism to choose representatives for concurrently facilitating the convergence and diversity of the proposed algorithm. In summary, based on the proposed nadir estimation method, the proposed dominance comparison approach, rank value and proximity distance assignment, and selection mechanism collectively improve the evolution of the proposed algorithm towards the PF with promising diversity. In order to qualify the performance of the proposed algorithm, a series of well-designed experiments is performed over two widely used benchmark test suites with $8$-, $15$-, and $20$-objective, their results measured by the selected performance metric indicate that the proposed algorithm is with considerable competitiveness in solving many-objective optimization problems. In addition, we utilize the proposed algorithm to solve one real-world many-objective optimization problem, in which the satisfactory results demonstrate the superiority of the proposed algorithm. Moreover, experiments are performed by the proposed decomposition-based nadir point estimation method against a couple of competitors over three representative test problems (DTLZ1, DTLZ2, and WFG2) with challenging features in PF shapes and objective value scales, the experimental results reveal the satisfactory results obtained by the proposed nadir point estimation method. In near future, we will place our efforts mainly on two essential aspects 1) constructing more accurate PF with limited information priori to obtaining the Pareto-optimal solutions to improve the development of indicator-based algorithms which require the uniformly distributed reference points, and 2) extending the proposed algorithm to solve constrained many-objective optimization problems.

\IEEEpeerreviewmaketitle
\ifCLASSOPTIONcaptionsoff
  \newpage
\fi


\begin{IEEEbiography}[{\includegraphics[width=1in,height=1.25in,clip,keepaspectratio]{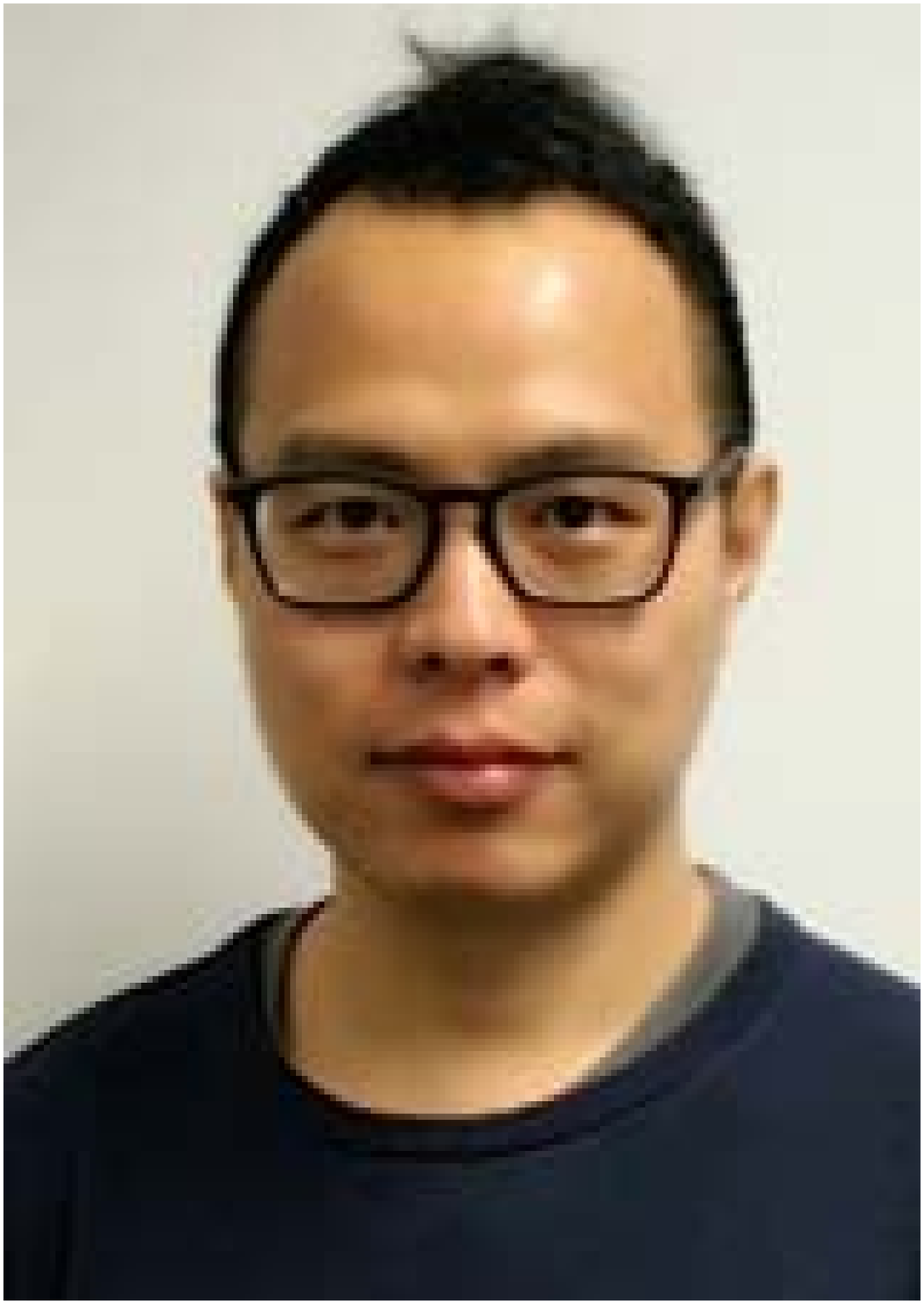}}]{Yanan Sun} (S'15-M'18) received a Ph.D. degree in engineering from the Sichuan University, Chengdu, China, in 2017. From 2015.08-2017.02, he is a jointly Ph.D. student financed by the China Scholarship Council in the School of Electrical and Computer Engineering, Oklahoma State University (OSU), USA. He is currently a Postdoc Research Fellow in the School of Engineering and Computer Science, Victoria University of Wellington, Wellington, New Zealand. His research topics are many-objective optimization and deep learning.
\end{IEEEbiography}

\begin{IEEEbiography}[{\includegraphics[width=1in,height=1.25in,clip,keepaspectratio]{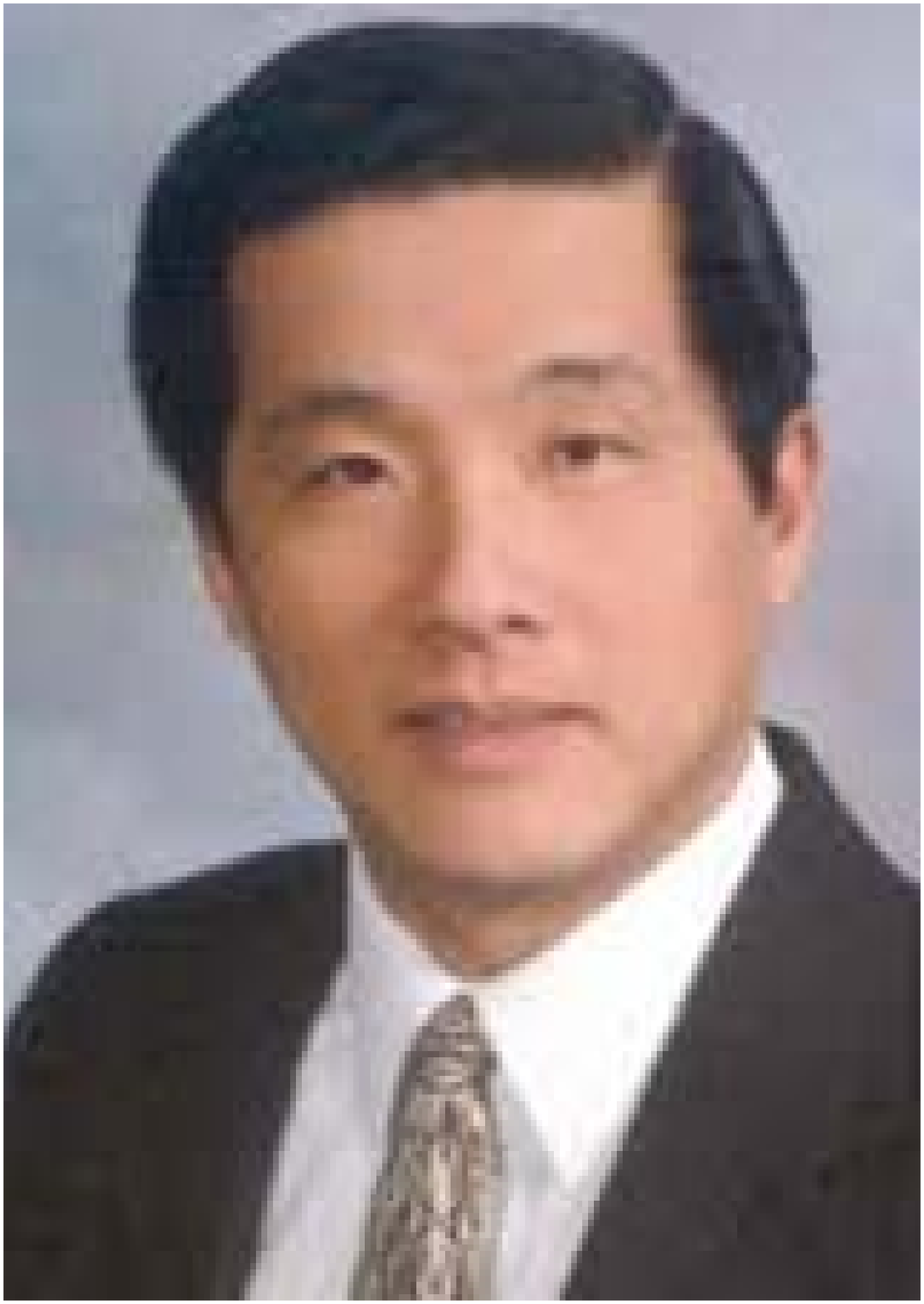}}]{Gary G. Yen}
(S'87-M'88-SM'97-F'09) received a Ph.D. degree in electrical and computer engineering from the University of Notre Dame in 1992. Currently he is a Regents Professor in the School of Electrical and Computer Engineering, Oklahoma State University (OSU). Before joined OSU in 1997, he was with the Structure Control Division, U.S. Air Force Research Laboratory in Albuquerque. His research interest includes intelligent control, computational intelligence, conditional health monitoring, signal processing and their industrial/defense applications.

Dr. Yen was an associate editor of the \textit{IEEE Control Systems Magazine, IEEE Transactions on Control Systems Technology}, \textit{Automatica}, \textit{Mechantronics}, \textit{IEEE Transactions on Systems, Man and Cybernetics, Parts A and B} and I\textit{EEE Transactions on Neural Networks}. He is currently serving as an associate editor for the \textit{IEEE Transactions on Evolutionary Computation} and the \textit{IEEE Transactions on Cybernetics}. He served as the General Chair for the \textit{2003 IEEE International Symposium on Intelligent Control} held in Houston, TX and \textit{2006 IEEE World Congress on Computational Intelligence} held in Vancouver, Canada. Dr. Yen served as Vice President for the Technical Activities in 2005-2006 and then President in 2010-2011 of the IEEE Computational intelligence Society. He was the founding editor-in-chief of the \textit{IEEE Computational Intelligence Magazine}, 2006-2009. In 2011, he received Andrew P Sage Best Transactions Paper award from \textit{IEEE Systems, Man and Cybernetics Society} and in 2014, he received Meritorious Service award from \textit{IEEE Computational Intelligence Society}.
\end{IEEEbiography}

\begin{IEEEbiography}[{\includegraphics[width=1in,height=1.25in,clip,keepaspectratio]{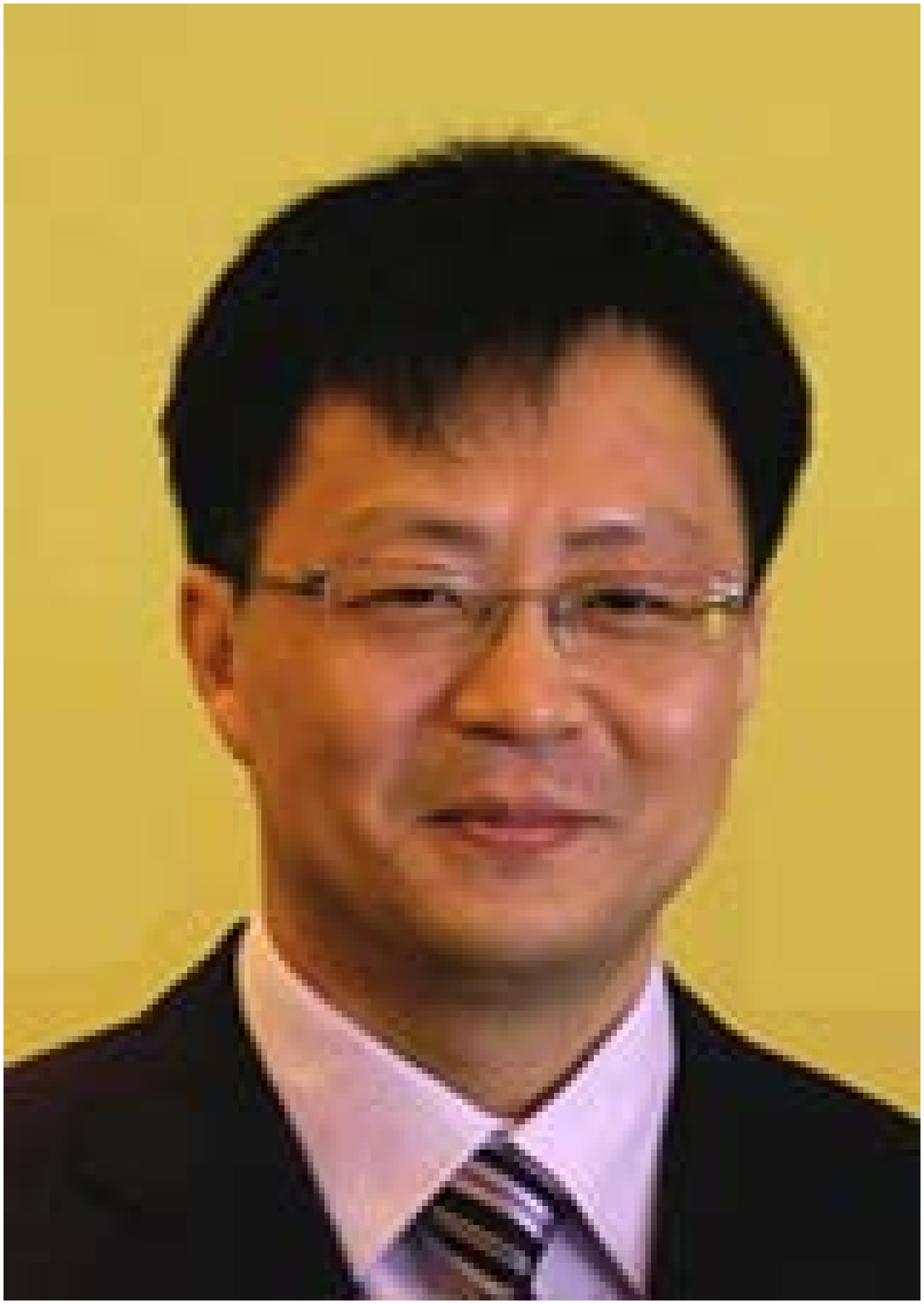}}]{Zhang Yi} (F'16)
received a Ph.D. degree in mathematics from the Institute of Mathematics, The Chinese Academy of Science, Beijing, China, in 1994. Currently, he is a Professor at the Machine Intelligence Laboratory, College of Computer Science, Sichuan University, Chengdu, China. He is the co-author of three books: \emph{Convergence Analysis of Recurrent Neural Networks} (Kluwer Academic Publishers, 2004), \emph{Neural Networks: Computational Models and Applications} (Springer, 2007), and \emph{Subspace Learning of Neural Networks} (CRC Press, 2010). He was an Associate Editor of \emph{IEEE Transactions on Neural Networks and Learning Systems} (2009~2012), and He is an Associate Editor of \emph{IEEE Transactions on Cybernetics} (2014~). His current research interests include Neural Networks and Big Data. He is a fellow of IEEE.
\end{IEEEbiography}
\newpage

\end{document}